\documentclass[10pt,letterpaper]{article}
\usepackage[top=0.85in,left=2.75in,footskip=0.75in,marginparwidth=2in]{geometry}
\usepackage[utf8]{inputenc}
\usepackage[version=3]{mhchem}
\usepackage{amsmath}
\usepackage{array}
\usepackage{booktabs}
\usepackage{cite}
\usepackage{float}
\usepackage{graphicx}
\usepackage{hyperref}
\usepackage{lastpage,fancyhdr}
\usepackage{lineno}
\usepackage{microtype}
\usepackage{needspace}
\usepackage{verbatim}
\usepackage{xcolor}
\usepackage[aboveskip=1pt,labelfont=bf,labelsep=period,singlelinecheck=off]{caption}

\graphicspath{{PDF.Fig/}}
\DisableLigatures[f]{encoding = *, family = * }
\raggedright
\fancyheadoffset[L]{2.25in}
\fancyfootoffset[L]{2.25in}

\makeatletter
\renewcommand{\@biblabel}[1]{\quad#1.}
\makeatother

\begin{document}
\vspace*{0.35in}

\begin{flushleft}
{\Large\textbf{Traceable Knowledge Graph Reasoning Enables LLM-Assisted Decision Support for Industrial VOCs in the Steel Industry}}
\newline\newline
Changqing Su\textsuperscript{1},
Yu Ding\textsuperscript{1},
Zuhong Lin\textsuperscript{1,*},
Hongyu Chen\textsuperscript{1},
Hongyu Liu\textsuperscript{1},
Xi He\textsuperscript{2},
Zheng Zeng\textsuperscript{3},
Liqing Li\textsuperscript{3,*}
\\
\bigskip
\textbf{1} Hunan Key Laboratory of Carbon Neutrality and Intelligent Energy, School of New Energy and Environment, Hunan University of Technology and Business, Changsha 410205, P.R.~China
\\
\textbf{2} Aerospace Kaitian Environmental Technology Co., Ltd., Changsha 410315, P.R.~China
\\
\textbf{3} School of Energy Science and Engineering, Central South University, Changsha 410083, P.R.~China
\\
\bigskip
* Corresponding authors: lzhzzzzwill@hutb.edu.cn; liqingli@hotail.com
\\
\end{flushleft}

\begin{abstract}
Key knowledge for steel-industry volatile organic compounds (VOCs) governance is scattered across unstructured scientific literature, making it difficult to integrate process, pollutant, and control-technology evidence and increasing the risk of hallucination when general large language models (LLMs) answer low-frequency industrial questions. Here we developed Chat-ISV, a knowledge graph (KG) enhanced multi-agent Q \& A system that parses a curated steel-industry VOCs literature corpus, constructs a Neo4j KG with 27180 nodes and 81779 semantic edges, and combines prompt-constrained extraction, chunk-centered topology optimization, multi-agent routing, source-backtracking retrieval, local literature retrieval, open-domain knowledge access, and interactive subgraph visualization. Benchmark tests and 400 expert blind evaluations showed that topology optimization reduced isolated nodes from 57\% to 4.08\% and that Chat-ISV achieved high factual reliability, with 96.93\% precision, 72.63\% recall, an F1-score of 0.830, and a mean score of 1.69/2.00. By converting fragmented environmental-engineering literature into traceable, queryable, and decision-support-oriented knowledge, Chat-ISV establishes a scalable environmental-informatics paradigm for reliable LLM deployment and intelligent pollution-control decision support in specialized industrial domains.

%\added[id=zh]{new}
%\deleted[id=zh]{delete}
%\replaced[id=zh]{new}{old}
%\comment[id=zh]{}
\end{abstract}

\section{Introduction}
The steel industry is a major anthropogenic source of volatile organic compounds (VOCs), with emissions that affect regional air quality, human health, downstream product-quality control, and pollution--carbon-reduction synergies\cite{Tang2022EnvRes,Liu2022JEM}. These coupled environmental and process-quality concerns make steel-industry VOC management a process-integrated challenge rather than an isolated end-of-pipe control task\cite{Lin2026LowCarbonScheduling}.
Extensive studies have examined steel-industry VOC emission inventories, synergistic reduction technologies, catalytic carbon conversion, and data-driven adsorbent design\cite{Zhang2022SynergisticControl,Chen2024STOTEN,Zhang2016ChemRevVOC,Zhou2025MethaneFeAl,Lin2025MOFArsenateGCN}. However, the resulting evidence on emission processes, VOC species, and control technologies remains dispersed across unstructured literature, limiting cross-source synthesis and targeted mitigation. Addressing this fragmented knowledge landscape requires dynamic, AI-assisted decision-support platforms built on integrated and continuously updated evidence streams\cite{Zhang2019CoalPolicy}.

Recent advances in large language models (LLMs) have opened new opportunities for mining high-entropy scientific literature\cite{Sun2026TII}. Prior studies have shown that LLMs can effectively support large-scale text understanding, tool-assisted chemical reasoning, and information extraction from massive corpora\cite{Miret2025CSR,Zheng2023ChatGPTChem,Duan2025LLMempowered,Kumar2025ALarge,White2023FutureChem,Taylor2022Galactica,Ouyang2022InstructGPT,Bran2024ChemCrow,Lin2026MOFMultiAgent}. Recent expert-system perspectives also emphasize that cross-disciplinary LLMs require explicit grounding in real-world domain evidence to remain reliable\cite{Zheng2026ExpertSystems}. In vertical domains, insufficient domain-specific prior knowledge nevertheless remains the major limitation. Wu et al.\cite{Wu2025PromptHallucination} systematically showed that general-purpose LLMs can produce chemically implausible predictions for key molecular properties such as topological polar surface area, whereas Zhou et al.\cite{Zhou2024CausalKGPT} reported that LLMs may generate unsupported causal chains in industrial fault diagnosis when explicit physical constraints are absent. These findings are consistent with broader surveys showing that factual hallucination remains a central obstacle to reliable deployment of LLMs in specialized scientific tasks\cite{Ji2023HallucinationSurvey,Huang2023HallucinationSurvey2}.

To mitigate factual hallucinations in vertical-domain LLM applications, recent works have introduced external-knowledge injection strategies, particularly retrieval-augmented generation (RAG) and knowledge graphs (KGs), to improve factual grounding and reasoning reliability\cite{Pan2024TKDE,Zhu2024LLMsfor,Agrawal2024Patterns,Gao2023RAGSurvey,Li2025TKDE}. Structured external knowledge has been shown to strengthen factual grounding and reasoning consistency in LLM-based systems, while representative design patterns for retrieval-enhanced generation further clarify the architectural basis for improving controllability, knowledge integration, and task adaptability in such systems\cite{Pan2024TKDE,Zhu2024LLMsfor,Agrawal2024Patterns}. Compared with conventional text-chunk retrieval, KG+LLM frameworks can explicitly organize dispersed evidence as entities and relations, support multi-hop retrieval, and provide clearer factual constraints\cite{Jafarzadeh2024AKG}. However, He et al.\cite{He2024GRetriever} showed that direct graph linearization can introduce context-window pressure and noise interference, Edge et al.\cite{Edge2024GraphRAG} demonstrated that overly local top-$k$ retrieval can weaken global connectivity and cross-subgraph reasoning, and Sun et al.\cite{Sun2024ToG,Xu2023Vulnerability} pointed out that sparse or fragmented graph topology can reintroduce hallucinations by interrupting the reasoning path available to the model. Building on these identified limitations, we hypothesized that, for VOCs governance in the steel industry, coordinated optimization of graph topology and query routing could reduce such risks more effectively by preserving global connectivity while maintaining explicit factual constraints.

\begin{figure}[htbp]
  \centering
  \includegraphics[width=1\textwidth]{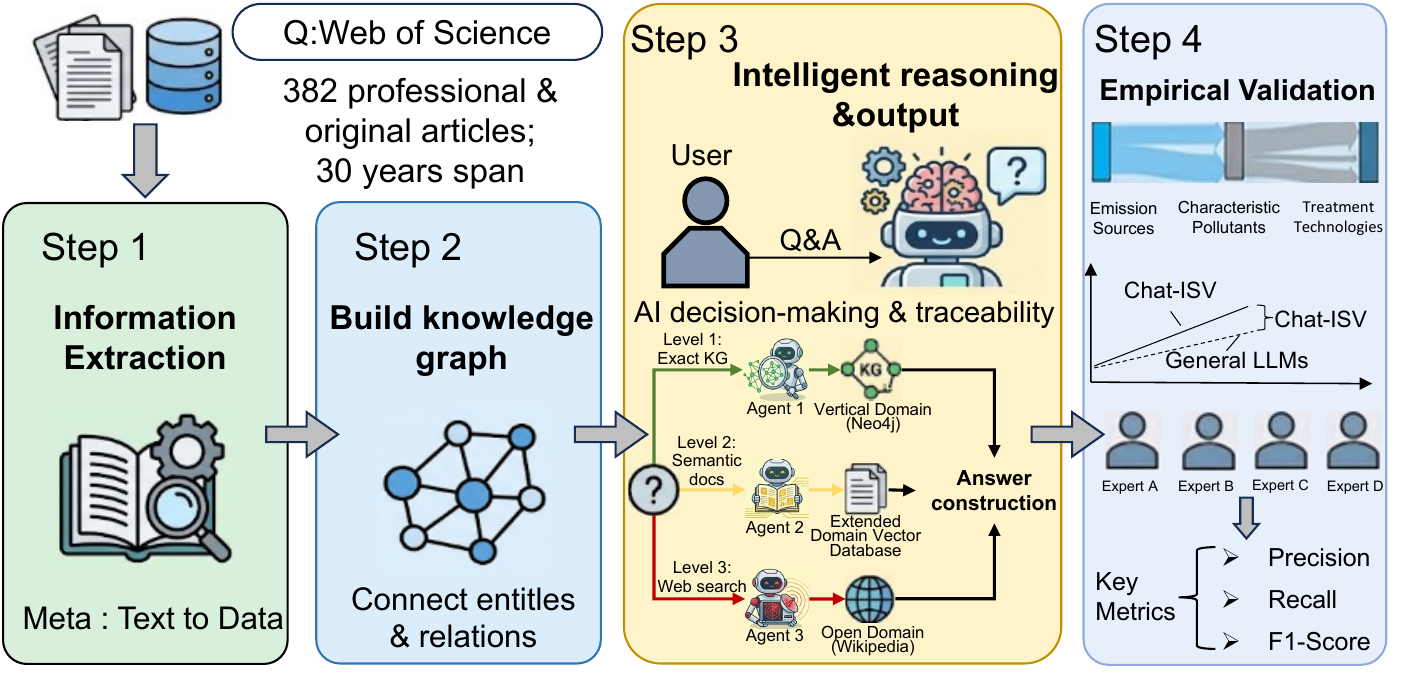}
  \caption{Pipeline of the Chat-ISV system. Unstructured knowledge from domain-specific scientific literature is extracted to construct a structured KG, and graph-retrieved evidence is incorporated into the LLM for question answering on VOCs governance in the steel industry.}
  \label{fig:pipeline}
\end{figure}

Accordingly, we developed Chat-ISV, a multi-agent collaborative question-answering system for VOCs governance in the steel industry (\textbf{Figure~\ref{fig:pipeline}}), following broader evidence that autonomous and conversational LLM agents can coordinate specialized reasoning roles in complex tasks\cite{Wang2024AgentSurvey,Wu2023AutoGen,Zhang2024NetworkedMultiagent}. The system is structured as an LLM-driven multi-agent KG framework that integrates automated knowledge extraction, topology optimization, and graph-guided query routing over 382 papers published over the past 30 years. In response to the topology- and retrieval-related limitations identified above, we combined front-end prompt intervention, back-end literature-chunk reconstruction, and multi-agent query routing to preserve graph connectivity while maintaining explicit factual constraints. On this basis, Chat-ISV translates natural-language questions into graph queries, supports multi-hop reasoning and source-text backtracking, and enables coordinated use of the vertical-domain KG and open-domain knowledge. We evaluate Chat-ISV by first examining how fragmented literature is transformed into structured steel-industry VOCs knowledge, then analyzing topology-enhanced traceable retrieval and topology-induced reasoning, and finally assessing factual reliability through cross-model benchmarking and expert blind evaluation.

\section{Methodology}

\subsection{Data Acquisition and Information Extraction}

We used the Web of Science (WoS) database as the primary source to systematically retrieve journal articles on VOCs emissions and control in the steel industry. The search strategy is detailed in \textbf{Text~S1} and \textbf{Table~S1} of the \textbf{Supporting Information} (\textbf{SI}). After screening for data completeness and extraction accuracy, a corpus of 382 full-text articles and supporting-information PDFs spanning nearly 30 years (1996-2025) was obtained.

Given the technical complexity of steel-industry VOCs-control literature, we developed an automated LLM-based pipeline to transform unstructured texts into structured knowledge. Each document was segmented into $\leq$2000-character chunks using a sliding-window strategy to ensure stable large-scale processing (\textbf{Algorithm~S1}).
We then constructed a domain-specific ontology comprising 9 entity types and 12 relation types, covering production processes, emission sources, VOCs species, control technologies, analytical methods, and contextual factors. Both direct physical relations (e.g., emission, control, measurement) and higher-level semantic associations were defined (\textbf{Table~S2}).
Within this ontology, the control-technology entity encompasses broad industrial processes as well as specific material-level interventions. For instance, the physical and chemical interactions between representative VOC species, including acetone and benzene, and advanced carbon-based adsorbents constitute critical decision nodes\cite{Su2020MaterChemPhys,Su2020Colloids}. By explicitly mapping these material properties and adsorption mechanisms into the graph, Chat-ISV links microscopic material-science innovations with macroscopic industrial application guidelines.

To enhance extraction accuracy, we implemented prompt engineering and a dual-track extraction mechanism. In the first track, we extracted explicit relations only when the text contained clear lexical or syntactic triggers. In the second track, we inferred implicit relations under constrained heuristics based on local co-occurrence and cross-sentence causal patterns. We also required source evidence and confidence scores for every extracted entity and relation to ensure traceability.

\subsection{Data Cleaning and Alignment}

Although LLMs enable effective unstructured information extraction, batch processing often introduces format inconsistencies and semantic drift. To ensure high usability of the resulting KG, we designed a dual-track data-governance pipeline integrating structural repair and semantic normalization.
At the structural level, malformed outputs were removed through streaming parsing and validation to preserve graph integrity. At the semantic level, entity ambiguity was resolved via normalization that separated canonical names from textual aliases and merged equivalent entities.

This two-layer strategy is designed to reduce redundancy and improve both structural robustness and semantic consistency, providing a reliable foundation for downstream graph-based retrieval and multi-hop reasoning (detailed automated extraction and normalization workflows are provided in \textbf{Text~S1.4}).

\subsection{Knowledge Graph Construction and Topology Optimization}
After data cleaning and normalization, we built a dedicated Cypher transaction script. The script efficiently mapped the standardized JSONL dataset into the Neo4j graph database (\textbf{Algorithm~S2}); the KG-construction JSONL dataset has been deposited in Zenodo with a persistent DOI\cite{Ding2026ChatISVDataset}. However, bottom-up LLM-extracted raw knowledge networks usually contained many isolated nodes without relation edges. To quantify graph connectivity and denoising effects, we used the isolated-node ratio as the core topological evaluation metric (\textbf{Equation~\ref{eqn:isolated_ratio}}).
\begin{equation}
P_{isolated}=\frac{N_{degree=0}}{N_{total}}\times 100\%
\label{eqn:isolated_ratio}
\end{equation}
where $N_{degree=0}$ denotes the number of isolated nodes that do not establish valid semantic relations with entities in other domains, and $N_{total}$ denotes the total number of entity nodes.

To improve graph connectivity, we combined front-end prompt intervention with back-end schema reconstruction. During extraction, semantic co-occurrence rules were added to the LLM prompt so that heuristic links could be inferred when explicit logical connectives were absent.

During data import, literature chunks were introduced as intermediate nodes in the Neo4j architecture. The system traced entities through the \texttt{evidence\_span} and \texttt{evidence\_text} attributes and created \texttt{:MENTIONS} edges from chunk nodes to entity nodes, yielding a chunk-centered star topology. Because chunk nodes may contain multiple entity types, this design accepts limited loss of local relation specificity in exchange for improved connectivity and source-level traceability.

\subsection{Knowledge-Graph Question-Answering System Design}
To move from static structured knowledge to dynamic intelligent interaction while addressing the factual hallucinations that large language models often produce in specialized domains, we developed and deployed a steel-industry VOCs question-answering system based on a hierarchical multi-agent collaborative architecture. The system integrates three complementary strengths: the explicit factual boundaries of the domain KG, the rich contextual detail preserved in local literature, and the broad reasoning capacity of large language models. Its core design is reflected in two closely connected dimensions: the hierarchical construction of multi-agent routing for closed-loop question answering and the optimization of the underlying graph-retrieval and source-tracing mechanism.

The Q \& A system adopts a three-tier routing framework comprising a graph-reasoning agent, a literature-retrieval agent, and an open-domain knowledge agent, coordinated by a scheduling agent. The scheduler identifies query intent, selects the retrieval path, merges returned evidence, and passes the assembled context to the base large language model for response generation (\textbf{Algorithm~S3}).

Retrieval over the vertical-domain KG serves as the primary route and uses an explicit node-anchoring strategy\cite{Hogan2021KG,Lewis2020RAG,Pan2024TKDE,Bai2025npjCompMat,Wang2026ChatRFB,Xu2024ChatTfA,Wang2025LLMKGMQA}. The system first loads graph-schema metadata as contextual guidance and then compiles natural-language questions into graph-query statements, allowing multi-hop traversal over direct and indirect relations. When explicit graph relations are insufficient, retrieval falls back to the local literature vector database and, if necessary, to open-domain resources.

Retrieved evidence is injected into the prompt template under source-backtracking constraints. During graph traversal, the retrieval algorithm backtracks from matched entities to tightly connected literature chunk nodes. These chunk nodes anchor domain entities such as production processes, pollutant species, and monitoring methods through \texttt{:MENTIONS} edges, as shown in \textbf{Figure~S1}. This design links graph-level relations to the underlying literature chunks and allows the retrieved context to include both structured entity relations and their corresponding source-text spans.

\subsection{Evaluation Metrics}
To objectively evaluate the generation quality and practical utility of the proposed model in realistic scenarios, we conducted a rigorous expert blind-evaluation work. An independent review panel of four experts with strong academic and industrial backgrounds evaluated 100 representative vertical-domain questions under a systematic double-blind protocol. Experts assigned scores across three core dimensions: scientific accuracy, logical rigor, and domain utility. Scientific accuracy assessed whether outputs conformed to objective domain facts and remained free of factual errors; logical rigor assessed coherence of reasoning under complex dependencies; and domain utility assessed direct applicability of outputs to practical research and engineering deployment.

To enable strict quantitative assessment, we mapped the expert score matrix (0--2) to a standard information-retrieval framework. A score of 2 was treated as a true positive (TP), indicating that the generated answer was fully accurate and completely traceable to underlying evidence. A score of 1 was treated as a false negative (FN), indicating that the answer was largely correct but omitted specific long-tail key entities. A score of 0 was treated as a false positive (FP), indicating severe factual error caused by deviation from the physical constraints of the underlying graph. A representative sample of the raw expert blind-evaluation dataset is provided in \textbf{Table~S6}, and the detailed scoring rubric is described in \textbf{Text~S3.2}. Under this mapping, Precision directly reflects the lower-bound capability to suppress factual errors, Recall captures completeness of knowledge extraction from the graph, and the F1-score reflects overall answer quality (the strict mathematical definitions of these metrics are detailed in \textbf{Text~S3.4}).

To further test whether the system architecture could generalize beyond the steel-industry VOCs corpus, we constructed an extended evaluation corpus from more than 4000 segmented documents in the Wikimedia Wikipedia dataset (\url{https://huggingface.co/datasets/wikimedia/wikipedia}). This auxiliary corpus covered broader environmental-engineering topics, including water treatment, marine pollution, carbon neutrality, and industrial mechanisms beyond steel manufacturing. The same four-expert blind-evaluation protocol and 0--2 scoring framework were then applied to the generated answers, producing another 400 valid ratings for generalization analysis (\textbf{Figure~S4}).

To evaluate downstream performance under a transparent quantitative scheme, we paired this retrieval framework with explicitly traceable outputs and a standardized expert-scoring matrix. To characterize overall answer quality while avoiding conclusions based solely on qualitative impressions, we calculated the expected score of the discrete expert ratings using a weighted-average formulation (\textbf{Equation~\ref{eqn:weighted_score}}):
\begin{equation}
\bar{S}=\frac{\sum_{k=0}^{2}k\cdot f_k}{N}
\label{eqn:weighted_score}
\end{equation}
where $k$ denotes the score level, $f_k$ denotes the frequency at score level $k$, and $N$ denotes the total number of ratings. To further assess whether the observed scoring pattern reflected stable expert judgment, we quantified inter-rater consistency using the Pearson correlation coefficient (\textbf{Equation~\ref{eqn:pearson}}):
\begin{equation}
r_{xy}=\frac{\sum_{i=1}^{n}(x_i-\bar{x})(y_i-\bar{y})}{\sqrt{\sum_{i=1}^{n}(x_i-\bar{x})^2}\sqrt{\sum_{i=1}^{n}(y_i-\bar{y})^2}}
\label{eqn:pearson}
\end{equation}
where $x_i$ and $y_i$ denote the scores assigned by two experts to sample $i$, $\bar{x}$ and $\bar{y}$ are the corresponding expert means, and $n$ is the number of evaluated samples. Together, these formulations provide the statistical basis for the performance and consistency analyses discussed below.

\section{Results and discussion}
\subsection{From Fragmented Literature to Structured Knowledge Insights into Steel-Industry VOCs Governance}
Through systematic extraction and topological denoising of more than 380 papers, we constructed a steel-industry VOCs governance KG containing 27180 nodes and 81779 semantic relation edges. This graph provides a structured basis for moving from fragmented bibliographic records to interpretable knowledge flows across sources, pollutants, and control technologies (\textbf{Figure~S2}).

\begin{figure}[htbp]
  \centering
  \includegraphics[width=1\textwidth]{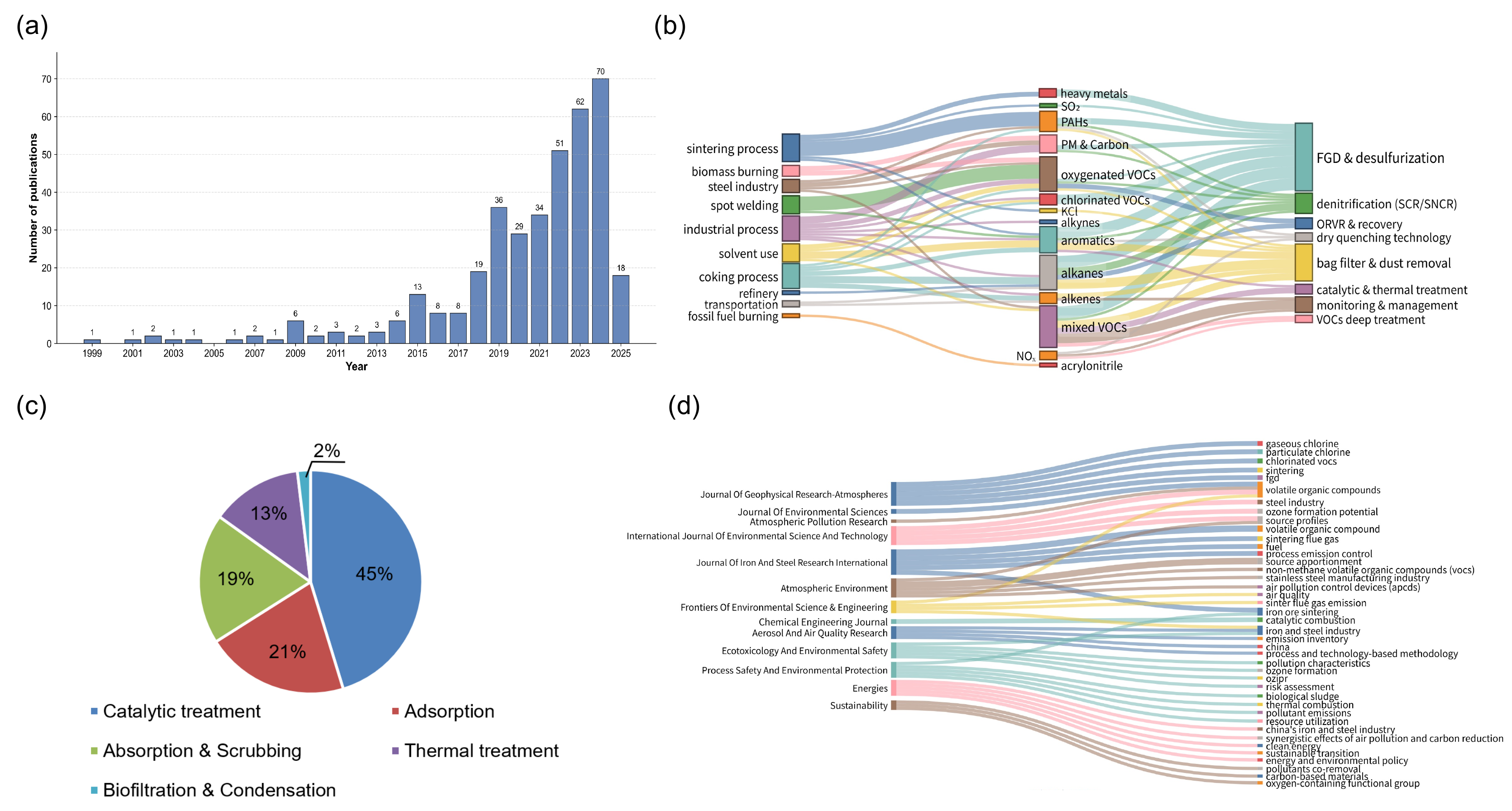}
  \caption{Bibliometric analysis and multidimensional knowledge-flow evolution in the steel-industry VOCs field. Overall publication growth trend in this field from 2008 to 2024 (a). Three-level Sankey diagram of emission sources, pollutant species, and control technologies reconstructed through multi-hop reasoning over the KG (b). Distribution of literature mention frequencies and relative shares of core deep-treatment technologies for VOCs control (c). Planar co-occurrence Sankey diagram exported directly from the traditional Web of Science platform (d).}
  \label{fig:bib}
\end{figure}

The publication trajectory first provides the temporal background for this knowledge system. As summarized in \textbf{Figure~\ref{fig:bib}a}, scholarly attention remained limited before 2015, but entered a rapid growth phase between 2018 and 2024, with annual publications increasing from 19 to 70. This acceleration indicates that VOCs governance in the steel industry has shifted from a peripheral environmental topic to an increasingly active research frontier driven by stricter emission-control demands and expanding process-level evidence.

Based on this expanding literature base, the KG-reconstructed Sankey diagram further translates dispersed textual evidence into a cross-layer knowledge-flow structure (\textbf{Figure~\ref{fig:bib}b}). On the source side, sintering, coking, general industrial processes, and spot welding emerge as major VOCs-related emission contexts, and long-tail sources such as solvent use and biomass burning are also retained by the graph reconstruction. The pollutant layer then reveals process-specific compositional patterns: sintering is more strongly associated with polycyclic aromatic hydrocarbons, whereas coking shows closer links to alkanes, aromatics, and alkenes. These pollutant groups ultimately converge on flue-gas desulfurization, denitrification, and bag-filter dust-removal units, suggesting that steel-industry VOCs mitigation is often embedded within coordinated multi-pollutant control infrastructure.

The same graph-derived evidence also clarifies the technological hierarchy of deep VOCs treatment. As shown in \textbf{Figure~\ref{fig:bib}c}, catalytic treatment accounts for 45.28\% of literature mentions, followed by adsorption separation (20.75\%), absorption scrubbing (18.87\%), and thermal combustion (13.21\%). Biological treatment and condensation recovery together account for less than 2\%. This distribution reflects a technology spectrum shaped by the high-temperature, large-volume, and compositionally complex exhaust streams typical of steel manufacturing, where catalytic and adsorption-based strategies currently dominate the research landscape.

The value of this KG-based reconstruction becomes clearer in comparison with the conventional Web of Science co-occurrence Sankey diagram (\textbf{Figure~\ref{fig:bib}d}). Because the traditional diagram depends primarily on literal text matching, it preserves keyword proximity with limited engineering interpretability. The graph-based representation in \textbf{Figure~\ref{fig:bib}b} aligns synonymous entities, resolves semantic ambiguity, and reconstructs directional relations among sources, pollutants, and control routes. This semantic structure provides a more coherent overview of the field and may support cautious hypothesis generation.

\subsection{Topology-Enhanced Knowledge Flow for Traceable Retrieval}
\begin{figure}[htbp]
  \centering
  \includegraphics[width=1\textwidth]{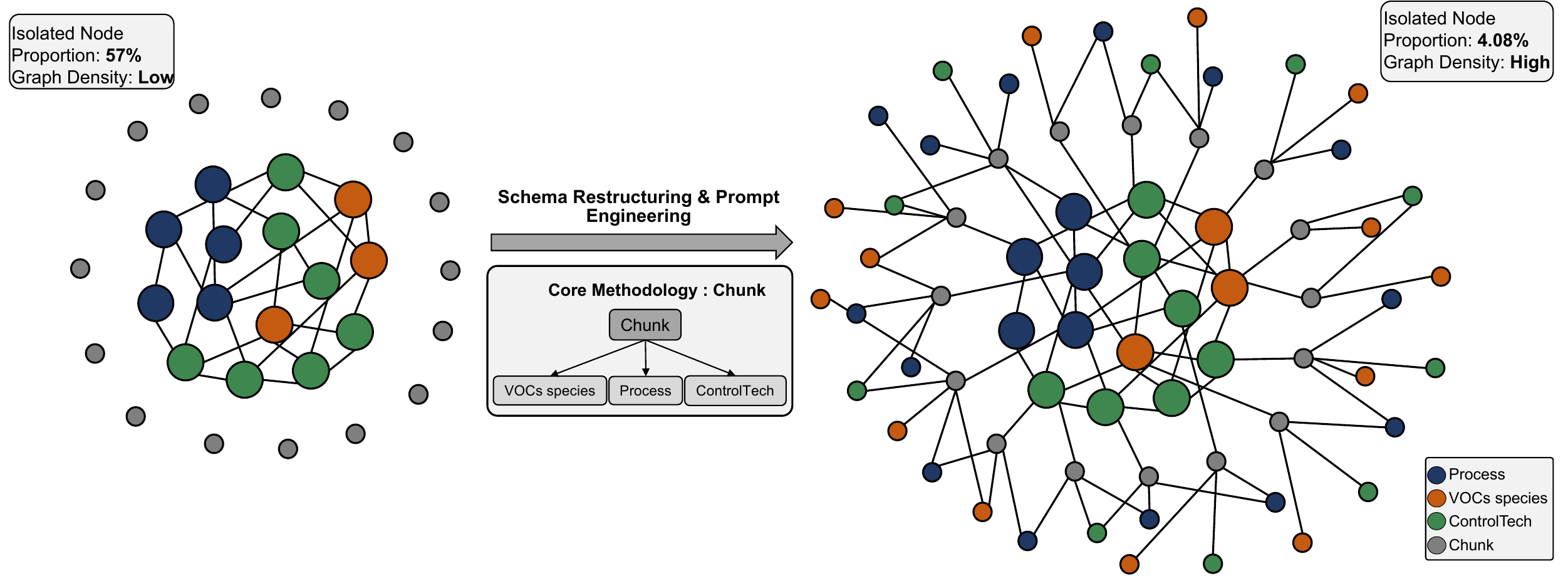}
  \caption{Local comparison of graph topology optimization.}
  \label{fig:optim}
\end{figure}

We used pilot optimization to compare successive topology-repair strategies following the two-stage procedure in \textbf{Algorithm~S4}. In the initial graph built only from basic LLM extraction, the isolated-node ratio was 57\%. In the first stage, front-end semantic completion supplemented explicit triples with domain-rule-based inferred triples and paragraph-level co-occurrence edges for entities that would otherwise remain isolated. This prompt- and rule-level intervention reduced the isolated-node ratio to 49\%, indicating improved connectivity but still insufficient support for complex knowledge retrieval. In the second stage, back-end schema restructuring introduced source chunk nodes, normalized entity aliases, created chunk--entity traceability edges, migrated long-tail observations into attributes, and removed redundant isolated nodes. After this chunk-level grounding step, the isolated-node ratio of the full graph decreased to 4.08\%. \textbf{Figure~\ref{fig:optim}} presents the resulting structural differences before and after topology optimization, with quantitative details summarized in \textbf{Text~S2} and \textbf{Tables~S3 and S4}.

We therefore attribute the main topological improvement to the combined effect of semantic completion and chunk-centered grounding. By reconnecting weakly linked entities to their source text through chunk nodes and \texttt{:MENTIONS} edges, this step established a complete traceability path from atomic nodes to original literature evidence.
We also observed an engineering trade-off. After adding chunk-level evidence links, part of the direct node-relation structure shifted from strictly semantic constraints toward context co-occurrence modeling. Because a single chunk node may contain multiple entity types, local relation specificity decreased slightly. In return, the optimized topology provided higher retrieval recall and better source-level traceability for downstream question answering (\textbf{Figure~S1}). The strategy of bridging discrete entities with chunk-level source nodes restructures the graph from a sparse semantic network into a denser evidence-grounded framework. This design is consistent with recent KG--LLM roadmaps and multi-document graph-prompting works showing that source-text grounding can improve logical coherence and traceability in downstream inference\cite{Pan2024TKDE,Wang2024KGPrompting}. It also addresses a recognized bottleneck in KG acquisition: automatically extracted graphs often suffer from sparsity, incompleteness, and weak connectivity, requiring topology-aware schema repair before robust graph reasoning can be performed\cite{Hogan2021KG,Ji2022KGSurvey,Sun2024ToG}.

\subsection{Topology-Induced Reasoning in the Steel-Industry VOCs Knowledge Space}
\begin{figure}[htbp]
  \centering
  \includegraphics[width=1\textwidth]{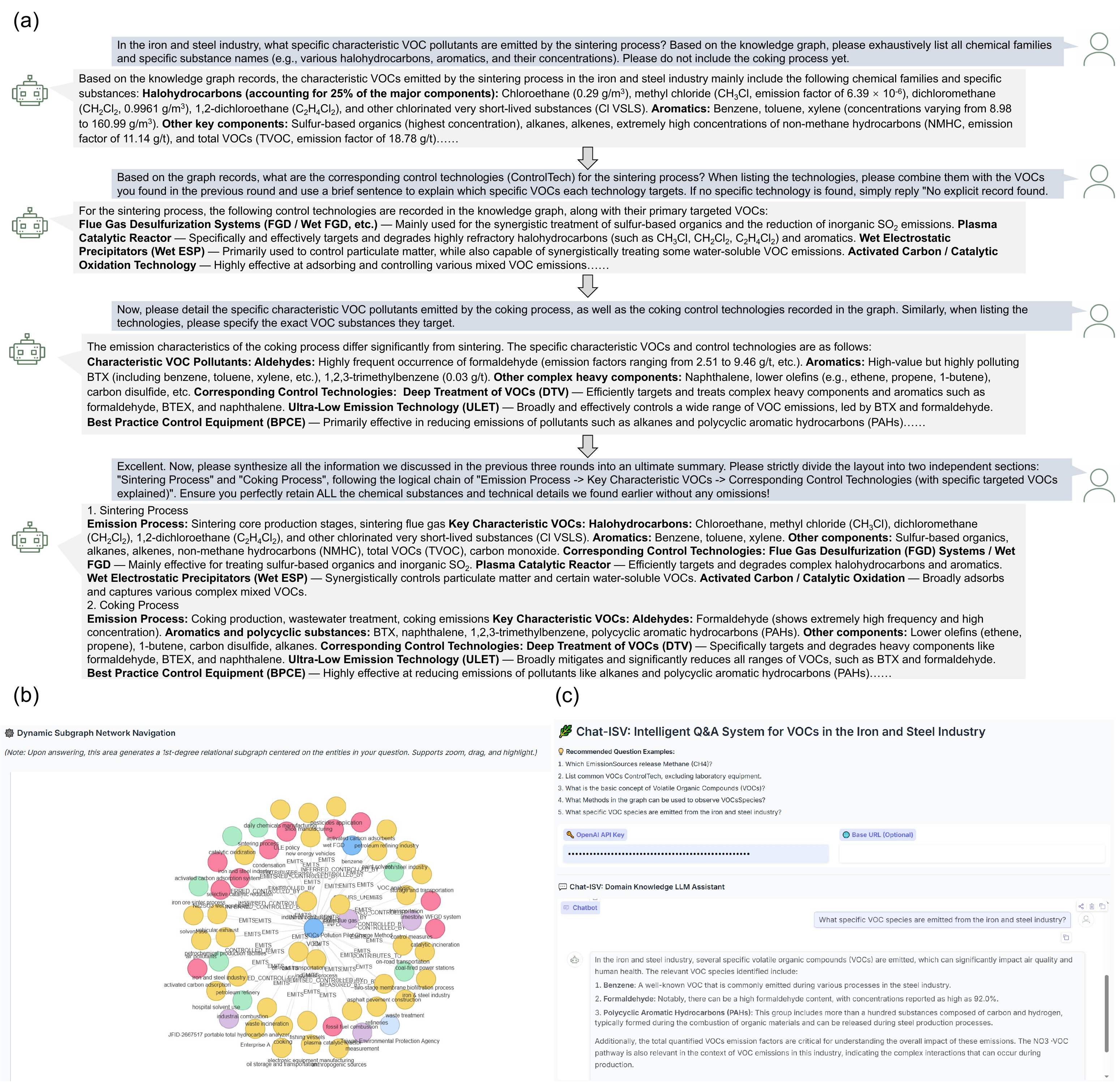}
  \caption{Intelligent question-answering interface and subgraph visualization: A heuristic multi-turn question answering workflow for the extraction and structured synthesis of the logical chain from emission processes to key characteristic VOCs and corresponding control technologies (a); Subgraph visualization of the KG (b); The Chat-ISV question-answering interface (c).}
  \label{fig:chat}
\end{figure}

Building on the topology-enhanced traceability established above, we next examined the reasoning role of the optimized KG topology. Specifically, we tested whether this topology could induce stepwise reasoning in the highly vertical steel-industry VOCs domain through constrained graph-grounded interactions. As illustrated in \textbf{Figure~\ref{fig:chat}b, c}, the workflow combines subgraph visualization with progressive user interaction. The displayed subgraph exposes the entities and relations linking emission processes, characteristic VOCs, control technologies, and literature evidence, allowing users to inspect the topological reasoning path before answer synthesis. The query process then unfolds sequentially to address the characteristic VOCs of the sintering process, the matched control technologies and their specific target pollutants, and the parallel data for the coking process. In the final interaction round, the instruction model uses its contextual memory to synthesize the previous interactions into a rigorous three-tier logical chain comprising the emission process, key characteristic VOCs, and corresponding control technologies. Step-by-step guidance preserves fine-grained evidence and prompts the model to explicitly map the asymmetric correspondence between technologies and specific pollutants, consistent with reasoning--acting paradigms that improve controllable intermediate reasoning in language-model workflows\cite{Yao2023ReAct}, thereby unlocking the topological value of the underlying KG.

This implementation also connects the above case work to the broader development of KG--LLM systems. Integrating KGs with large language models has been explored across multiple disciplines as a way to improve response quality in vertical domains\cite{Pan2024TKDE,Dreger2025DigitDiscov,Miret2025CSR}. Building on this research direction, Chat-ISV further emphasizes practical accessibility for environmental-engineering researchers. As shown in \textbf{Figure~\ref{fig:chat}a}, we provide a zero-code, user-friendly web interface at \url{https://huggingface.co/spaces/Grace-Ding/VOCs-graph}. Through this interface, domain scholars can interactively query the VOCs-governance knowledge base and explore the Chat-ISV service without coding expertise.

\begin{figure}[htbp]
  \centering
  \includegraphics[width=1\textwidth]{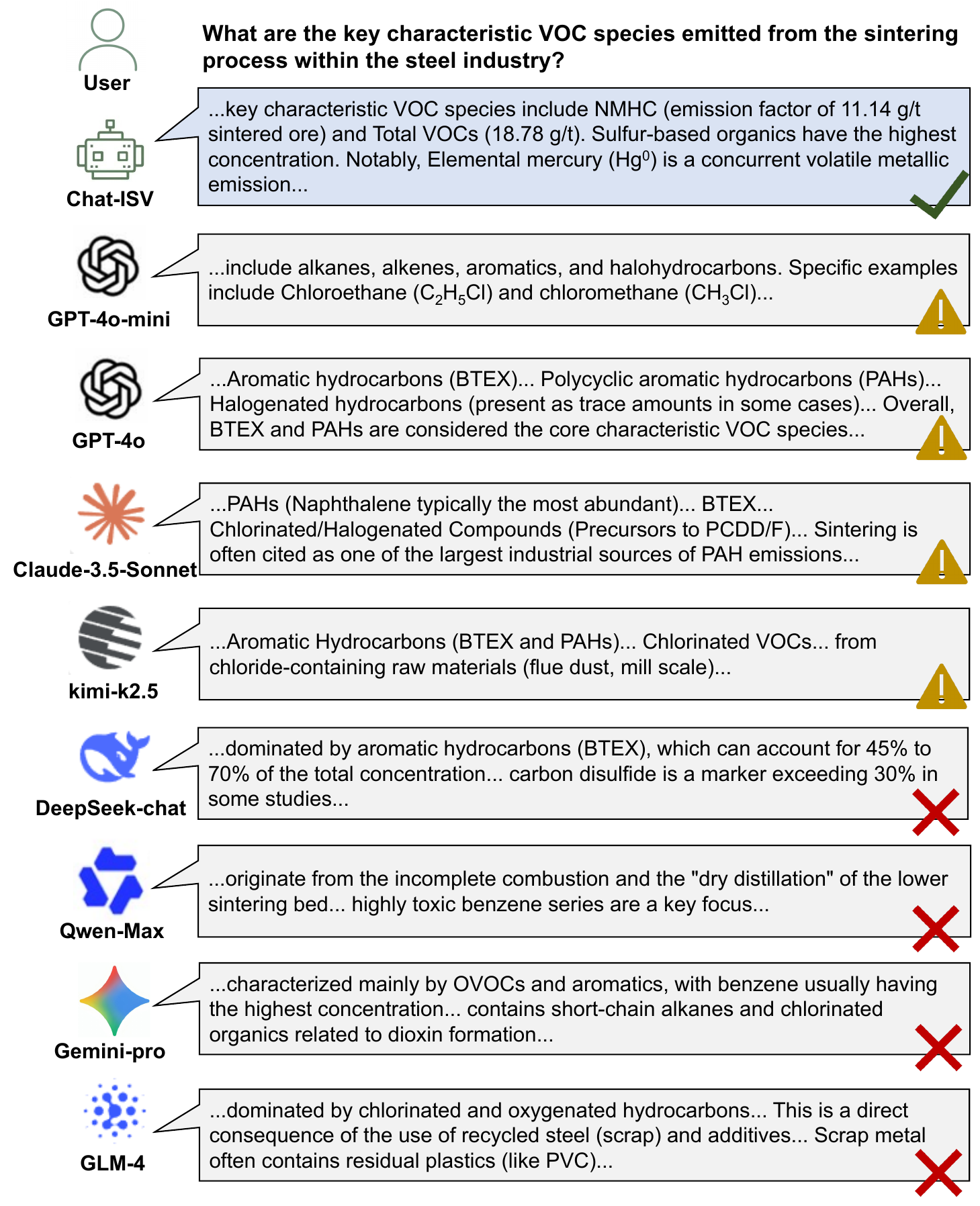}
  \caption{Comprehensive cross-model performance evaluation.}
  \label{fig:compare}
\end{figure}

After establishing this user-facing reasoning workflow, we further benchmarked Chat-ISV against commercial general-purpose LLMs. The comparison is summarized in \textbf{Figure~\ref{fig:compare}} (\textbf{Text~S3.1 and Table~S5}). For a highly specialized sintering-VOCs query, the general models exhibited different forms of cognitive limitation and factual hallucination.
In this horizontal evaluation, Claude-3.5-Sonnet and Kimi-k2.5 produced coherent but generic environmental-chemistry summaries. They missed process-specific evidence, such as unusually high concentrations of long-tail chlorinated species in sintering emissions, and did not provide accurate industrial emission factors. GPT-4o showed a plausible-but-generic bias by transferring coking-related characteristics, such as benzene-series compounds, to the sintering process.

Other models produced more explicit process-level hallucinations. GLM-4 attributed chlorinated VOCs to PVC plastics in scrap steel, thereby confusing sintering based on iron-ore fines with electric-arc-furnace steelmaking. Gemini-pro instead identified oxygenated organics or benzene as dominant species, which is inconsistent with industrial sintering conditions. These cases show that, without precise domain constraints, fluent commercial LLMs can generate persuasive but misleading answers.

Chat-ISV performs topological traversal and precise matching on the structured KG. It retrieves literature chunks containing explicit evidence---for example, 11.14 g of non-methane hydrocarbons per ton of sinter and chloromethane at 0.9973 mg/m$^3$---and injects these verified passages as context. This hard-constraint retrieval keeps answers grounded in traceable source data, supports component-level evidence recovery for long-tail pollutants, and reduces numerical and factual hallucinations at the information-access stage. Generative models are particularly prone to hallucinations when dealing with highly specific scientific parameters, such as the exact enhancement effects of nitrogen or phosphorus doping on VOC adsorption capacities\cite{Su2025CJCHE,Su2020MaterChemPhys}. By grounding the reasoning process through source-linked literature chunks and \texttt{:MENTIONS} edges, Chat-ISV ensures that sensitive engineering metrics and material-design parameters are retrieved from peer-reviewed literature rather than generated probabilistically. Because generative hallucinations remain persistent in knowledge-intensive and high-stakes domains, Chat-ISV uses multi-agent cascade routing as an architectural intervention rather than relying only on prompt wording\cite{Ji2023HallucinationSurvey,Talebirad2023MultiAgent}. By decoupling graph traversal, semantic vector recall, and final response generation, the system isolates factual retrieval from generative synthesis, aligning with retrieval-augmented and agent-society findings that dynamic retrieval and multi-agent interaction can improve domain-specific answer grounding\cite{Lewis2020RAG,Lan2024AgentSociety}.

Overall, this comparison shows that the entity-text joint KG improves large-model reasoning by combining a macro-level knowledge skeleton with micro-level empirical evidence. By anchoring answers to validated graph topology and source text, Chat-ISV compensates for the weak domain recall of general-purpose LLMs and provides a more reliable basis for narrow vertical-domain research and engineering applications.

\subsection{Factual Reliability of Chat-ISV in Steel-Industry VOCs Question Answering}
\begin{figure}[H]
  \centering
  \includegraphics[width=1\textwidth]{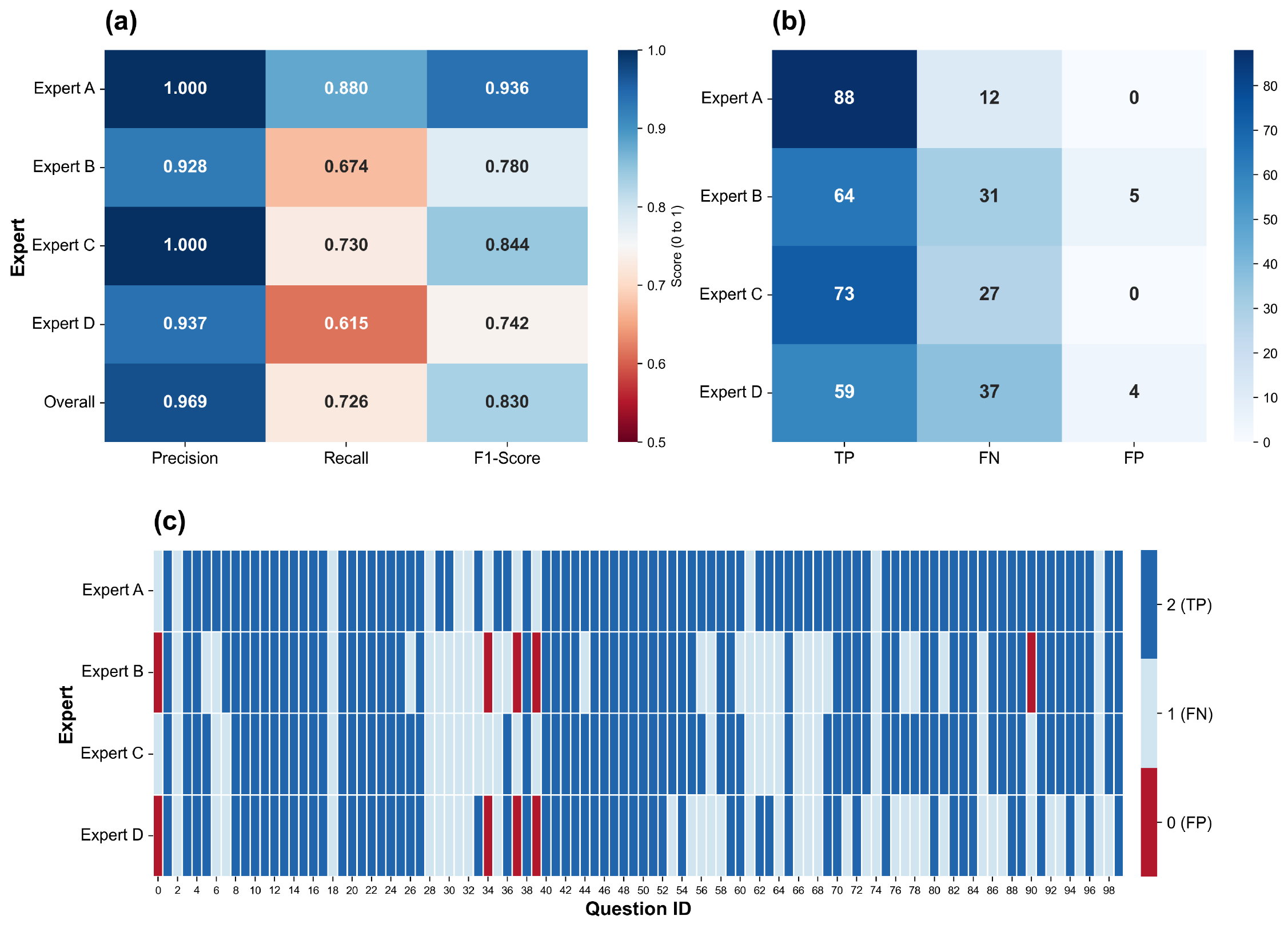}
  \caption{Comprehensive quantitative evaluation and consistency analysis of the Q \& A system based on expert blind assessment: Heatmap of Precision, Recall, and F1-score evaluated independently by four experts (a); Count-distribution matrix of true positives (TP, 2 points), false negatives (FN, 1 point), and false positives (FP, 0 points) assigned by the review panel (b); Panoramic expert-consistency map recording individual scoring variations across 100 vertical-domain test questions (c).}
  \label{fig:performance}
\end{figure}

Chat-ISV was developed to provide factually reliable question answering for steel-industry VOCs governance, where incorrect pollutant--process--control associations may mislead environmental diagnosis and engineering decisions. As shown in \textbf{Figure~\ref{fig:performance}a, b}, Chat-ISV showed strong factual reliability in the steel-industry VOCs Q \& A benchmark. Across 400 independent blind ratings, the system received 284 TP ratings, 107 FN ratings, and only 9 FP ratings, corresponding to an overall precision of 96.93\%, recall of 72.63\%, and F1-score of 0.830. Experts A and C assigned no zero-point ratings in their respective evaluations, further indicating that the traceable KG topology effectively constrained divergence-driven fabrication. The moderate recall reduction mainly occurred in extreme long-tail multi-hop queries, where the routing algorithm favored high-confidence local subgraphs over low-confidence peripheral expansion. This high-precision, medium-to-high-recall profile is consistent with industrial environmental-safety scenarios that prioritize factual correctness.

The weighted-average score further reached 1.69 out of 2.00, confirming that most answers were judged as fully or largely correct. The expert-consistency map in \textbf{Figure~\ref{fig:performance}c} showed pairwise Pearson correlations of 0.57--0.77 among the four experts, with the highest consistency between Experts B and C (0.77; \textbf{Figure~S3}). These moderate-to-high correlations reflect broadly convergent expert judgment. Together, the high precision, low FP rate, and stable inter-expert consistency support the factual reliability of Chat-ISV for complex vertical-domain Q \& A tasks.

The extended-corpus test also demonstrated strong generality of the architecture (\textbf{Figure~S4}). Across 400 additional expert ratings, the system achieved a global precision of 91.79\%, with 241 fully correct ratings, 137 partially correct ratings, and only 22 severe-error ratings. The corresponding severe-error proportion was 5.5\%, indicating that the framework remained robust even outside the dedicated steel-industry VOCs KG.

Under this generalization setting, recall and F1-score reached 65.08\% and 76.18\%, respectively. The lower recall mainly reflects the system's conservative safety-isolation mechanism: when the external corpus lacked sufficient evidence, Chat-ISV preferred factually correct summary-level answers or mandatory exits within the retrieval boundary. These results indicate that the architecture retains retrieval-boundary awareness and low factual-error risk across broader scientific and industrial topics.

\section{Conclusion}
In this work, we developed Chat-ISV, a KG-enhanced LLM question-answering system for steel-industry VOCs control. Using a curated steel-industry VOCs literature corpus, we constructed a domain KG with 27180 nodes and 81779 semantic relation edges. By combining prompt-constrained extraction, chunk-centered topology reconstruction, multi-agent routing, and source-backtracking retrieval, Chat-ISV links emission sources, VOCs species, control technologies, and literature evidence within a traceable reasoning framework.
The resulting system achieved the intended goal of reducing hallucination and improving evidence-grounded answering in a low-frequency industrial domain. Topology optimization reduced the isolated-node ratio from 57\% in the initial LLM-extracted graph to 4.08\%, thereby improving graph connectivity and source-level traceability. In 400 expert blind evaluations, Chat-ISV achieved a precision of 96.93\%, a recall of 72.63\%, an F1-score of 0.830, and a mean expert score of 1.69 out of 2.00, with inter-expert Pearson correlations ranging from 0.57 to 0.77. In the extended generalization test, the system maintained a precision of 91.79\%, with recall and F1-score reaching 65.08\% and 76.18\%, respectively.
These results indicate that Chat-ISV can transform fragmented industrial-environmental literature into traceable, queryable, and decision-support-oriented knowledge, providing a practical route for reliable LLM application in specialized engineering domains. Its strong factual reliability and generalization potential further suggest that KG-enhanced LLM systems can serve as scalable infrastructure for intelligent pollution-control decision support. To further strengthen this capability, future work will focus on dynamic KG updates, broader cross-pollutant and cross-industry expansion, and integration with real-time monitoring data.

\section*{Acknowledgments}
This work was funded by the Project of Hunan Provincial Social Science Achievements Review Committee (XSP24YBZ060), the “Digital-intelligence+” interdisciplinary research project of Hunan University of Technology and Business (2023SZJ22) and Supported by the Postgraduate Research Innovation Project of Hunan University of Technology and Business (CX2025YB035 and CX2025YB034).

\section*{Supporting Information Statement}
Supporting Information is provided as a separate file, including supplementary methods, algorithms, evaluation details, figures, and tables.

\section*{Code Availability}
The source code and workflow implementation of Chat-ISV, including scripts for knowledge extraction, graph construction, retrieval, question answering, and visualization, are available in the project GitHub repository (\url{https://github.com/DYDY911/Chat-ISV/tree/main}).

\section*{Data Availability}
The KG-construction JSONL dataset is publicly available in the Zenodo repository (\url{https://doi.org/10.5281/zenodo.20280089}). The Q \& A experimental records are provided in the project repository text-record file (\url{https://github.com/DYDY911/Chat-ISV/blob/main/text_record/text.log}), and the expert-scored benchmark dataset is available in the project repository test-data folder (\url{https://github.com/DYDY911/Chat-ISV/tree/main/test_data}).

\section*{Declaration of generative AI and AI-assisted technologies in the writing process}
During the preparation of this work, the author(s) used Claude to assist in correcting grammatical errors in the text. After using this tool/service, the author(s) reviewed and edited the content as needed and take(s) full responsibility for the content of the published article.

\section*{Conflict of Interest}
The authors declare no competing financial interest.

\section*{Author Contributions}
C.S.: conceptualization, methodology, software, data curation, formal analysis, visualization, and writing---original draft. Y.D.: investigation, validation, data curation, and writing---review and editing. Z.L.: supervision, funding acquisition, project administration, conceptualization, and writing---review and editing. H.C. and H.L.: data checking, literature organization, and validation. X.H.: industrial-domain consultation and application-scenario validation. Z.Z. and L.L.: supervision, project guidance, and writing---review and editing. All authors reviewed and approved the final manuscript.
\clearpage
\section*{Supporting Information}
\addcontentsline{toc}{section}{Supporting Information}
\subsection*{Text S1 Data Acquisition and Ontology Specifications}
\addcontentsline{toc}{subsection}{Text S1 Data Acquisition and Ontology Specifications}

\subsubsection*{S1.1. Literature Collection and Corpus Construction}
To ensure both domain specificity and high data quality in the knowledge base, we conducted a systematic literature search using the Web of Science (WoS) Core Collection. Corpus construction followed a rigorous manual verification protocol.

\textbf{Search strategy.} As summarized in \textbf{Table~S1}, search queries covered macro-level emissions, source profiles, and key subprocesses, including sintering, coking, and ironmaking.

\textbf{Manual acquisition.} We identified 382 highly relevant publications (1996--2025). Each article was manually accessed and downloaded as a full-text PDF to avoid data loss associated with abstract-only parsing.

\textbf{Local repository construction.} All files were converted into machine-readable text using a customized Python pipeline (\texttt{llm\_extra\_pdf.py} in the code repository (\url{https://github.com/DYDY911/Chat-ISV})), while preserving domain-specific technical terminology and numerical values.

\subsubsection*{S1.2. Text Preprocessing and Segmentation}
Because of the context-window constraints of large language models (LLMs), we applied a sliding-window segmentation strategy to decompose long scientific documents into manageable units.

\textbf{Window size.} Each chunk was restricted to \(\leq\)2,000 characters.

\textbf{Semantic continuity.} We implemented controlled overlap between adjacent chunks to reduce boundary truncation of key entities, relations, and causal chains.

\textbf{Pipeline resilience.} A breakpoint-resume mechanism was integrated to support stable large-scale batch extraction and prevent duplicate processing.

\subsubsection*{S1.3. Domain Ontology and Schema Definition}
The knowledge graph (KG) is governed by a domain-specific ontology designed to capture complex interdependencies in industrial environmental engineering. \textbf{Table~S2} summarizes the ontology schema, including representative definitions of the nine entity categories and 12 relation types used in Chat-ISV.

\subsubsection*{S1.4. Automated Extraction and Normalization Workflow}
To transform unstructured literature into a Neo4j-ready graph representation, we implemented a dual-track governance workflow.

\textbf{Structural repair.} We used a heuristic bracket-counting algorithm to repair truncated JSON fragments generated by the LLM, ensuring robust downstream parsing for valid outputs.

\textbf{Entity normalization.} Synonyms and variant technical expressions (e.g., ``RTO'' vs ``regenerative thermal oxidizer'') were mapped to canonical forms through a dictionary-based normalization layer, thereby reducing node redundancy and improving graph consistency.

\clearpage
\subsection*{Text S2 Topological Synergistic Optimization Data}
\addcontentsline{toc}{subsection}{Text S2 Topological Synergistic Optimization Data}
During initial graph construction, the knowledge graph exhibited severe knowledge fragmentation. To meet the requirements of complex multi-hop retrieval in intelligent question-answering, we implemented a two-stage synergistic optimization strategy that combined front-end heuristic intervention with back-end schema restructuring.

\subsubsection*{S2.1. Initial Topological Bottleneck}
In the initial stage, which relied primarily on baseline LLM-based information extraction, the proportion of isolated nodes (degree = 0) reached approximately 57\%. As illustrated in \textbf{Figure~S3}, this low connectivity directly reflects severe fragmentation of the initial knowledge network. Such fragmentation substantially constrained macro-level analysis of knowledge flow and limited the execution of graph-walk-based retrieval algorithms, thereby motivating subsequent structural optimization.

\subsubsection*{S2.2. Stage 1: Front-End Heuristic Intervention}
To mitigate topological disconnection, Stage~1 introduced semantic co-occurrence reasoning rules into prompt engineering. Under this design, the model was allowed to create heuristic edges from sentence- or paragraph-level context even when explicit discourse connectives were absent. According to low-level quantitative statistics from the graph database, this intervention reduced the isolated-node ratio to 49.88\% (36,397 isolated nodes among 72,967 total nodes). Although connectivity improved, nearly half of the nodes remained disconnected, which was still insufficient for comprehensive recall in industrial-grade knowledge retrieval.

\subsubsection*{S2.3. Stage 2: Chunk-Level Traceable Topology Restructuring}
To fundamentally break the connectivity bottleneck and remove redundancy, the system implemented deep schema restructuring and entity normalization during Neo4j ingestion.

\textbf{Hub-node introduction.} We introduced \texttt{Chunk} as a new intermediate node type at the physical storage layer.

\textbf{Star-topology construction.} The system enforced \texttt{:MENTIONS} links from each \texttt{Chunk} node to domain entities.

\textbf{Redundancy collapsing.} Based on this traceable network, many synonym nodes and fragmented observation nodes generated in early extraction were consolidated and pruned.

After this back-end reconstruction, the isolated-node ratio dropped sharply and stabilized at 4.08\% (1,109 isolated nodes among 27,180 total nodes).

\subsubsection*{S2.4. Quantitative Statistics of Node and Relationship Distributions}
To further quantify the effects of two-stage topological optimization, \textbf{Tables~S3} and \textbf{S4} compare low-level node and relationship distributions between Stage~1 (post-prompt intervention) and Stage~2 (post-schema restructuring). The results indicate that the system substantially increased network density and traceable links while merging redundant entities.

\noindent $^{*}$\textit{Observation} nodes are long-tail numeric attachment nodes generated during initial LLM extraction. In Stage~2, many values were converted into entity attributes or chunk attributes, substantially compressing redundant topology.

\clearpage
\subsection*{Text S3 Benchmarking and Expert Evaluation Dataset}
\addcontentsline{toc}{subsection}{Text S3 Benchmarking and Expert Evaluation Dataset}
To objectively evaluate the real-world performance of Chat-ISV in the steel-industry VOC domain, we designed a multi-model benchmarking experiment and a rigorous expert blind-evaluation protocol. This section provides the detailed experimental setup, representative question-answer comparison cases, and a sample of the expert-scoring dataset.

\subsubsection*{S3.1. Multi-Model Benchmarking}
To verify the necessity of domain knowledge graphs (KGs) for reducing factual hallucinations in large language models (LLMs), we conducted a horizontal comparison against four leading general-purpose pretrained models: GPT-4o-mini, DeepSeek-chat, Qwen-Max, and Gemini-pro.

All models were evaluated under identical conditions (temperature = 0.2, zero-shot prompting) on highly specialized industrial-environmental questions. \textbf{Table~S5} presents a representative case. The results indicate that, without bottom-layer KG constraints, general models tend to produce high-plausibility but factually unreliable outputs (e.g., fabricated dominant pollutant categories or omission of critical co-emitted pollutants). In contrast, Chat-ISV can retrieve quantitative industrial evidence with substantially higher fidelity.

\subsubsection*{S3.2. Expert Blind-Evaluation Design and Raw Dataset}
To quantify scientific accuracy, logical rigor, and domain utility of system outputs, we invited four domain experts with strong environmental-engineering backgrounds (Expert A, B, C, and D) to conduct blind evaluation.

\textbf{Sample set.} The full set includes 100 professional questions covering VOC emission inventories, synergistic control pathways, regulatory standards, and chemical mechanisms.

\textbf{Scoring rubric.}

\textbf{2 points (full score):} factually correct, logically coherent, evidence-traceable, and directly useful for engineering or research.

\textbf{1 point (moderate):} largely correct but lacking depth, missing key details, or showing mild contextual deviation.

\textbf{0 points (deficient):} clear factual errors, severe data hallucination, or major logical inconsistency.

\noindent \textit{Note:} To satisfy SI length constraints, \textbf{Table~S6} shows only representative samples with contrasting score patterns. The complete 100-question raw scoring matrix is provided as a separate Excel attachment.

\subsubsection*{S3.3. Inter-Rater Reliability Analysis}
As shown in \textbf{Figure~S4}, we extracted 400 ratings (100 questions $\times$ 4 experts) from the complete underlying expert-evaluation dataset. The weighted mean score of the discrete outcomes was 1.69. To rule out mechanically homogeneous scoring behavior, we further calculated Pearson correlation coefficients from pairwise expert score vectors $(X, Y)$.

Even when objective disagreements occurred for individual questions within the full evaluation set, the global inter-rater correlation matrix remained stably within 0.57--0.77. From a statistical perspective, this supports both the authenticity of the blind-evaluation process and the scientific validity of the evaluation framework.

\subsubsection*{S3.4. Definition and Calculation of Quantitative Performance Metrics}
To rigorously quantify the system's ability to suppress factual hallucinations and ensure knowledge completeness, we mapped the 0--2 expert scoring matrix to a standard information-retrieval evaluation framework using true positives (TP), false negatives (FN), and false positives (FP):

\textbf{True Positive (TP) = 2 points:} the model generates a fully accurate and logically rigorous answer with precise traceable evidence grounded in the underlying knowledge graph.

\textbf{False Negative (FN) = 1 point:} the answer is largely correct but misses specific long-tail entities or fails to sufficiently traverse the graph (i.e., correct knowledge that should have been retrieved is omitted).

\textbf{False Positive (FP) = 0 points:} the model departs from the physical constraints of the knowledge graph and generates severe factual errors or data hallucinations (i.e., incorrect knowledge is produced).

Based on this mapping, we calculated Precision, Recall, and F1-score for each independent expert evaluation and for overall system performance. Precision directly represents the lower-bound capability of the system to suppress factual hallucinations, Recall measures knowledge-extraction completeness, and F1-score reflects overall QA quality. The metric definitions are:

\begin{equation}
\mathrm{Precision}=\frac{TP}{TP+FP}
\end{equation}

\begin{equation}
\mathrm{Recall}=\frac{TP}{TP+FN}
\end{equation}

\begin{equation}
\mathrm{F1-score}=2\times\frac{\mathrm{Precision}\times\mathrm{Recall}}{\mathrm{Precision}+\mathrm{Recall}}
\end{equation}

To support the generalization-capability assessment, we conducted an independent evaluation using environmentally relevant documents extracted from the Wikipedia corpus. Four domain experts performed blind scoring of the system outputs on a 0--2 scale, yielding a total of 400 valid ratings. \textbf{Figure~S5} provides a detailed visualization of the score distributions assigned by all experts, together with the derived performance metrics, including Precision, Recall, and F1-score.

\clearpage
\subsection*{Algorithms}
\addcontentsline{toc}{subsection}{Algorithms}
To clearly illustrate the computational procedures and the multi-agent routing logic proposed in this study, the fundamental workflows of the Chat-ISV architecture are formalized in the following pseudocodes.

Algorithm~S1 details the automated pipeline for extracting unstructured PDF texts into structured graph triples using large language models (LLMs) with checkpointing mechanisms. Algorithm~S2 describes the ingestion process of the extracted JSON objects into the Neo4j graph database, establishing explicit links between entities and their source literature chunks. Algorithm~S3 outlines the core three-tier routing mechanism, which dynamically allocates user queries across the knowledge graph, the local vector database, and the open-domain Wikipedia API to reduce hallucination risk. Algorithm~S4 formalizes the two-stage graph-topology optimization procedure for high-fidelity traceability.

\Needspace{24\baselineskip}
\subsubsection*{Algorithm S1. Automated Knowledge-Graph Extraction via LLM}
\addcontentsline{toc}{subsubsection}{Algorithm S1. Automated Knowledge-Graph Extraction via LLM}
\textbf{Input:} Raw literature documents $\mathcal{D}$, chunk size $S$, prompt template $\mathcal{P}$, LLM parameters (temperature $\tau$, maximum retries $N_{\mathrm{retry}}$)\\
\textbf{Output:} Structured JSONL dataset containing entities and relations $\mathcal{J}$

{\footnotesize
\begin{tabular}{@{}r p{0.92\linewidth}@{}}
1: & Initialize an empty set for processed chunks $\mathcal{C}_{\mathrm{processed}} \leftarrow \emptyset$ \\
2: & Load existing extraction progress from checkpoint file to $\mathcal{C}_{\mathrm{processed}}$ \\
3: & \textbf{for} each document $D_i \in \mathcal{D}$ \textbf{do} \\
4: & \hspace*{1em}Extract raw text $T_i$ from $D_i$ \\
5: & \hspace*{1em}Split $T_i$ into textual chunks $\{c_{i,1}, c_{i,2}, \dots, c_{i,m}\}$ of size $S$ \\
6: & \hspace*{1em}\textbf{for} each chunk $c_{i,j}$ \textbf{do} \\
7: & \hspace*{2em}Generate unique chunk identifier $id \leftarrow \mathrm{Hash}(D_i, j)$ \\
8: & \hspace*{2em}\textbf{if} $id \in \mathcal{C}_{\mathrm{processed}}$ \textbf{then} \\
9: & \hspace*{3em}\textbf{continue} \\
10: & \hspace*{2em}\textbf{end if} \\
11: & \hspace*{2em}Format instruction $\mathcal{P}_{i,j}$ by inserting $c_{i,j}$ into prompt template $\mathcal{P}$ \\
12: & \hspace*{2em}Initialize retry counter $k \leftarrow 0$ \\
13: & \hspace*{2em}\textbf{while} $k < N_{\mathrm{retry}}$ \textbf{do} \\
14: & \hspace*{3em}$\mathcal{R}_{i,j} \leftarrow \mathrm{Call\_LLM}(\mathcal{P}_{i,j}, \tau)$ \\
15: & \hspace*{3em}\textbf{if} $\mathcal{R}_{i,j}$ matches the predefined JSON schema \textbf{then} \\
16: & \hspace*{4em}Append $\mathcal{R}_{i,j}$ to output file $\mathcal{J}$ \\
17: & \hspace*{4em}Add $id$ to $\mathcal{C}_{\mathrm{processed}}$ and update checkpoint \\
18: & \hspace*{4em}\textbf{break} \\
19: & \hspace*{3em}\textbf{else} \\
20: & \hspace*{4em}$k \leftarrow k + 1$ \\
21: & \hspace*{4em}Wait for exponential backoff time \\
22: & \hspace*{3em}\textbf{end if} \\
23: & \hspace*{2em}\textbf{end while} \\
24: & \hspace*{1em}\textbf{end for} \\
25: & \textbf{end for} \\
26: & \textbf{return} $\mathcal{J}$
\end{tabular}
}

\clearpage
\subsubsection*{Algorithm S2. Knowledge-Graph Ingestion and Entity--Source Grounding}
\addcontentsline{toc}{subsubsection}{Algorithm S2. Knowledge-Graph Ingestion and Entity--Source Grounding}
{\footnotesize
\setlength{\tabcolsep}{2pt}
\renewcommand{\arraystretch}{0.78}
\begin{tabular}{@{}r p{0.90\linewidth}@{}}
\multicolumn{2}{@{}p{0.98\linewidth}@{}}{\textbf{Input:} Extracted JSONL dataset $\mathcal{J}$, Neo4j database instance $\mathcal{G}$, batch size $B$}\\
\multicolumn{2}{@{}p{0.98\linewidth}@{}}{\textbf{Output:} Populated multi-relational knowledge graph $\mathcal{G}$}\\[-0.2em]
1: & Initialize memory buffer $\mathcal{B} \leftarrow \emptyset$ \\
2: & \textbf{for} each record $r \in \mathcal{J}$ \textbf{do} \\
3: & \hspace*{1em}Append $r$ to buffer $\mathcal{B}$ \\
4: & \hspace*{1em}\textbf{if} $\mathrm{Size}(\mathcal{B}) \geq B$ \textbf{then} \\
5: & \hspace*{2em}Begin graph transaction $Tx$ \\
6: & \hspace*{2em}\textbf{for} each item in $\mathcal{B}$ \textbf{do} \\
7: & \hspace*{3em}Extract $chunk\_id$, $doc\_id$, and $text$ from item \\
8: & \hspace*{3em}Create chunk node $N_C$ in $\mathcal{G}$ with properties $(chunk\_id, text)$ \\
9: & \hspace*{3em}\textit{Process entities} \\
10: & \hspace*{3em}\textbf{for} each category and entity list in item.entities \textbf{do} \\
11: & \hspace*{4em}\textbf{for} each entity $e$ \textbf{do} \\
12: & \hspace*{5em}Merge entity node $N_e$ (create if not exists) \\
13: & \hspace*{5em}Create relationship $N_e \xrightarrow{\mathrm{MENTIONED\_IN}} N_C$ \\
14: & \hspace*{4em}\textbf{end for} \\
15: & \hspace*{3em}\textbf{end for} \\
16: & \hspace*{3em}\textit{Process relations} \\
17: & \hspace*{3em}\textbf{for} each relation $rel \in$ item.relations \textbf{do} \\
18: & \hspace*{4em}Identify source node $N_{\mathrm{head}}$ and target node $N_{\mathrm{tail}}$ \\
19: & \hspace*{4em}Merge relationship $N_{\mathrm{head}} \xrightarrow{rel.type} N_{\mathrm{tail}}$ \\
20: & \hspace*{4em}Set edge properties $(evidence\_text, confidence)$ \\
21: & \hspace*{3em}\textbf{end for} \\
22: & \hspace*{2em}\textbf{end for} \\
23: & \hspace*{2em}Commit transaction $Tx$ \\
24: & \hspace*{2em}Clear buffer $\mathcal{B} \leftarrow \emptyset$ \\
25: & \hspace*{1em}\textbf{end if} \\
26: & \textbf{end for}
\end{tabular}
}

\clearpage
\subsubsection*{Algorithm S3. Three-Tier Multi-Agent Question-Answering and Evidence Routing}
\addcontentsline{toc}{subsubsection}{Algorithm S3. Three-Tier Multi-Agent Question-Answering and Evidence Routing}
{\footnotesize
\setlength{\tabcolsep}{2pt}
\renewcommand{\arraystretch}{0.86}
\begin{tabular}{@{}r p{0.90\linewidth}@{}}
\multicolumn{2}{@{}p{0.98\linewidth}@{}}{\textbf{Input:} User query $Q$, Neo4j knowledge graph $\mathcal{G}$, FAISS vector database $\mathcal{V}$, Wikipedia API $\mathcal{W}$, evaluation rules $R_{\mathrm{eval}}$}\\
\multicolumn{2}{@{}p{0.98\linewidth}@{}}{\textbf{Output:} Final answer $A$ with evidence trace $\mathcal{E}$}\\[-0.2em]
1: & Parse $Q$ to identify domain entities, relation intents, constraints, and expected answer type \\
2: & \textit{Tier 1: Knowledge-graph reasoning agent} \\
3: & Generate Cypher query $C$ from the parsed intent and ontology schema \\
4: & Execute $C$ on $\mathcal{G}$ to obtain graph evidence $\mathcal{E}_{\mathrm{kg}}$ \\
5: & \textbf{if} $\mathcal{E}_{\mathrm{kg}}$ is sufficient under $R_{\mathrm{eval}}$ \textbf{then} \\
6: & \hspace*{1em}$A \leftarrow \mathrm{Generate\_Answer}(Q,\mathcal{E}_{\mathrm{kg}})$ \\
7: & \hspace*{1em}\textbf{return} $(A,\mathcal{E}_{\mathrm{kg}})$ \\
8: & \textbf{end if} \\
9: & \textit{Tier 2: Local literature-retrieval agent} \\
10: & Retrieve top-$k$ semantically relevant chunks $\mathcal{E}_{\mathrm{local}}$ from $\mathcal{V}$ \\
11: & Rerank $\mathcal{E}_{\mathrm{local}}$ by entity overlap, process relevance, and evidence density \\
12: & \textbf{if} $\mathcal{E}_{\mathrm{local}}$ passes factual-grounding checks in $R_{\mathrm{eval}}$ \textbf{then} \\
13: & \hspace*{1em}$A \leftarrow \mathrm{Generate\_Answer}(Q,\mathcal{E}_{\mathrm{local}})$ \\
14: & \hspace*{1em}\textbf{return} $(A,\mathcal{E}_{\mathrm{local}})$ \\
15: & \textbf{end if} \\
16: & \textit{Tier 3: Open-domain fallback agent} \\
17: & Generate keyword query $Q_{\mathrm{wiki}}$ from unresolved entities and relation intents \\
18: & Retrieve open-domain context $\mathcal{E}_{\mathrm{wiki}}$ from $\mathcal{W}$ \\
19: & Generate conservative answer $A$ and mark unsupported domain-specific claims as unresolved \\
20: & Attach source tier, retrieved evidence, and fallback status to $\mathcal{E}$ \\
21: & \textbf{return} $(A,\mathcal{E}_{\mathrm{wiki}})$
\end{tabular}
}

\vspace{0.5em}
\Needspace{24\baselineskip}
\subsubsection*{Algorithm S4. Two-Stage Graph Topology Optimization and Traceability Enhancement}
\addcontentsline{toc}{subsubsection}{Algorithm S4. Two-Stage Graph Topology Optimization and Traceability Enhancement}
\textbf{Input:} Preliminary entity--relation graph $G_0=(V_0,E_0)$, extracted chunk set $\mathcal{C}$, normalization dictionary $D_{\mathrm{norm}}$, co-occurrence rule set $R_{\mathrm{co}}$\\
\textbf{Output:} Denoised and traceable knowledge graph $G^{*}=(V^{*},E^{*})$

{\footnotesize
\begin{tabular}{@{}r p{0.92\linewidth}@{}}
1: & Initialize optimized node set $V^{*} \leftarrow \emptyset$ and edge set $E^{*} \leftarrow \emptyset$ \\
2: & \textit{Stage 1: Front-end semantic completion} \\
3: & \textbf{for} each chunk $c \in \mathcal{C}$ \textbf{do} \\
4: & \hspace*{1em}Extract explicit triples $T_{\mathrm{exp}}$ from syntactic and semantic triggers in $c$ \\
5: & \hspace*{1em}Infer supplementary triples $T_{\mathrm{inf}}$ using domain rules and process constraints \\
6: & \hspace*{1em}\textbf{if} entities remain isolated in $c$ \textbf{then} \\
7: & \hspace*{2em}Generate paragraph-level co-occurrence edges $T_{\mathrm{co}}$ according to $R_{\mathrm{co}}$ \\
8: & \hspace*{1em}\textbf{else} set $T_{\mathrm{co}} \leftarrow \emptyset$ \\
9: & \hspace*{1em}Update $V^{*}$ and $E^{*}$ using $T_{\mathrm{exp}} \cup T_{\mathrm{inf}} \cup T_{\mathrm{co}}$ \\
10: & \textbf{end for} \\
11: & \textit{Stage 2: Back-end schema restructuring and grounding} \\
12: & \textbf{for} each chunk $c \in \mathcal{C}$ \textbf{do} \\
13: & \hspace*{1em}Create or merge source chunk node $N_c$ with raw-text and document metadata \\
14: & \hspace*{1em}\textbf{for} each entity node $N_e$ mentioned in $c$ \textbf{do} \\
15: & \hspace*{2em}Normalize $N_e$ to canonical form using $D_{\mathrm{norm}}$ \\
16: & \hspace*{2em}Create traceability edge $(N_e,\ \mathrm{MENTIONED\_IN},\ N_c)$ \\
17: & \hspace*{1em}\textbf{end for} \\
18: & \textbf{end for} \\
19: & Merge duplicate aliases, migrate long-tail observations to attributes, and remove redundant isolated nodes \\
20: & \textbf{return} $G^{*}=(V^{*},E^{*})$
\end{tabular}
}

\clearpage
\addcontentsline{toc}{subsection}{Figure S1. Chunk-Centered Star-Topology Local Schematic}
\begin{figure}[H]
  \centering
  \includegraphics[width=1\textwidth,height=0.78\textheight,keepaspectratio]{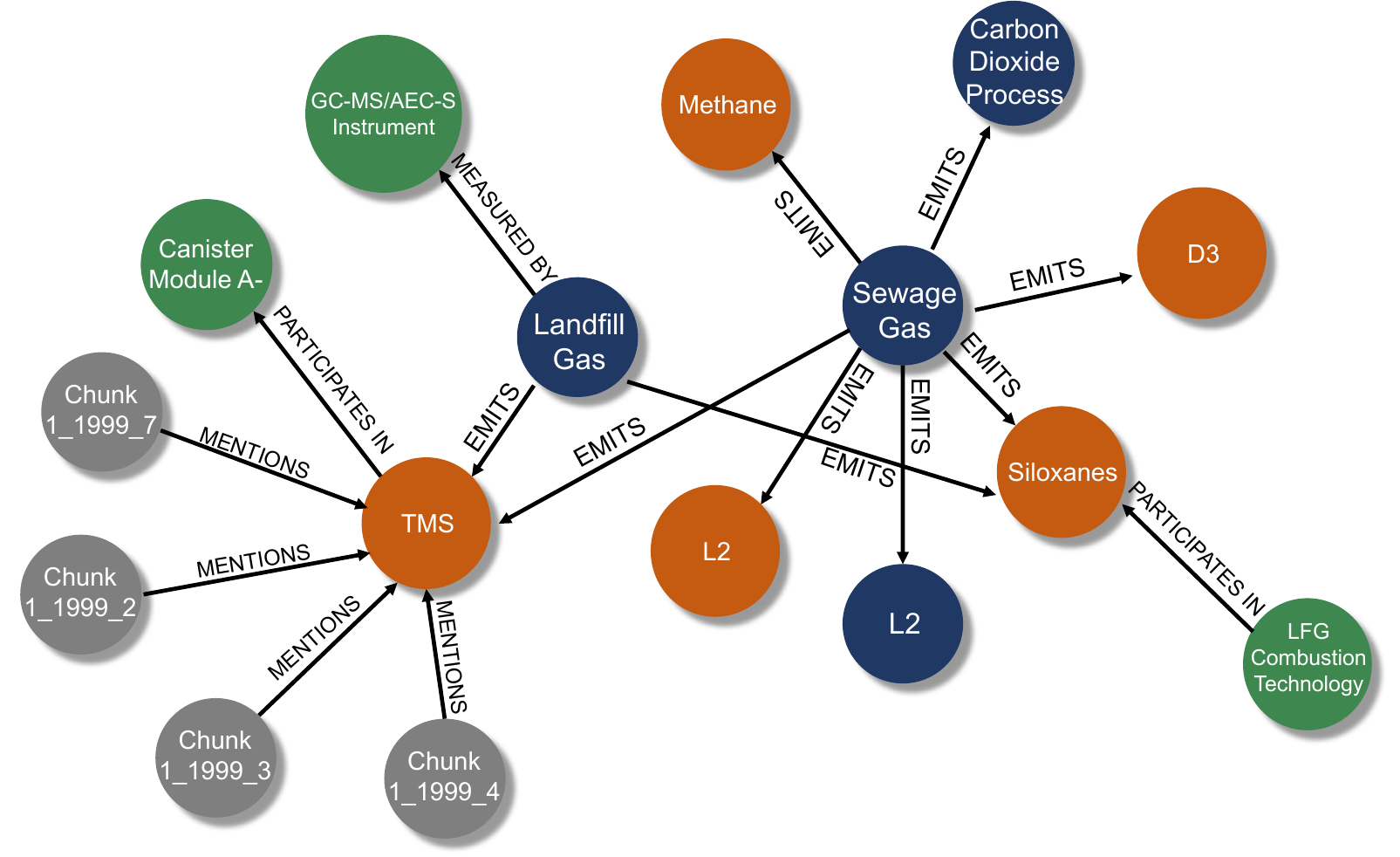}
  \caption{Chunk-centered star-topology local schematic in the Chat-ISV system.}
  \label{fig:s_topo}
\end{figure}

\clearpage
\addcontentsline{toc}{subsection}{Figure S2. Schema Visualization of the Optimized Knowledge Graph}
\begin{figure}[H]
  \centering
  \includegraphics[width=1\textwidth,height=0.78\textheight,keepaspectratio]{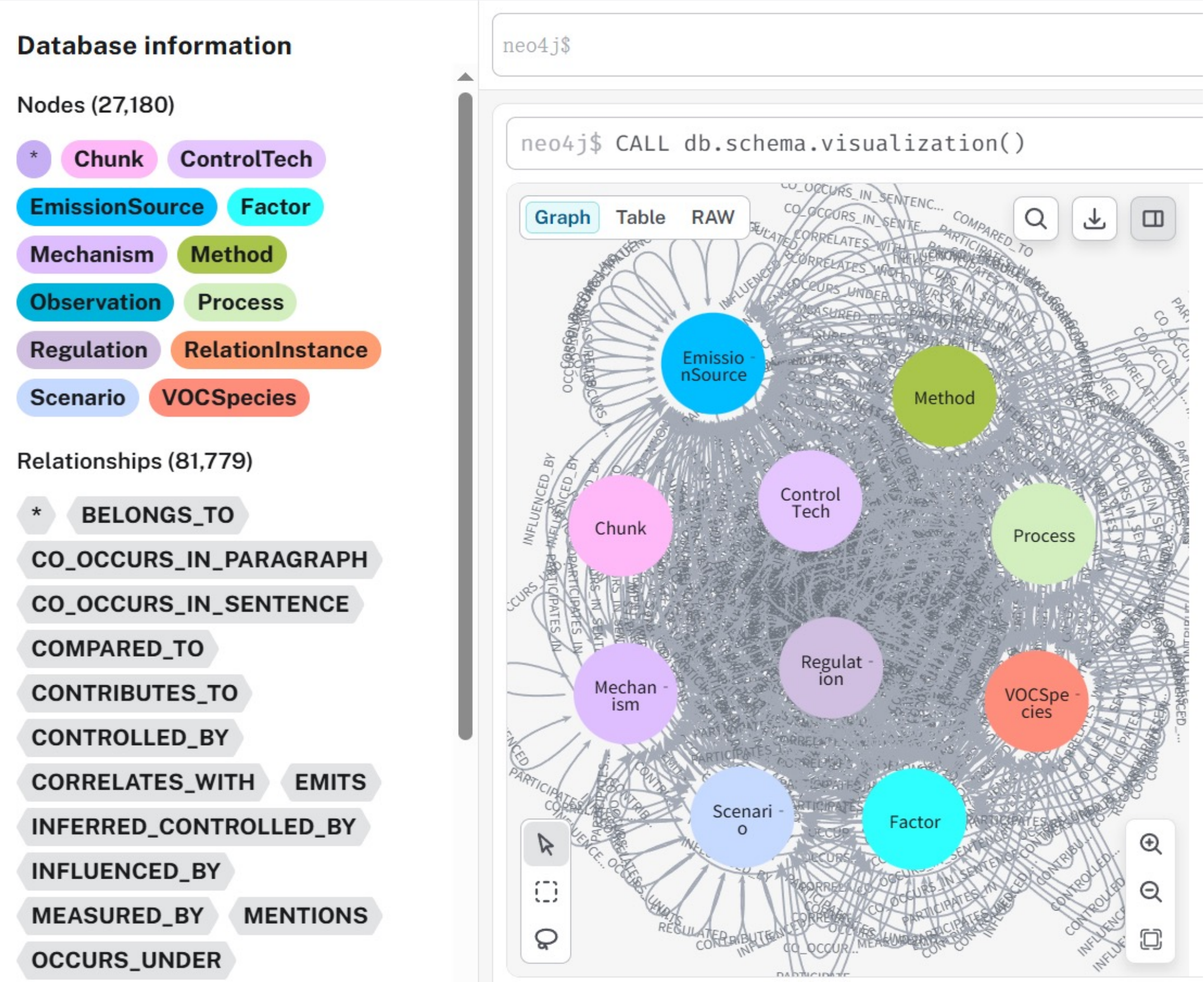}
  \caption{Schema visualization and database statistics of the optimized knowledge graph in Neo4j. The completed topology integrates 27180 entity nodes and 81779 semantic relationship edges.}
  \label{fig:s_allinfo}
\end{figure}

\clearpage
\addcontentsline{toc}{subsection}{Figure S3. Snapshot of the Initial Knowledge-Graph Topology}
\begin{figure}[H]
  \centering
  \includegraphics[width=1\textwidth,height=0.78\textheight,keepaspectratio]{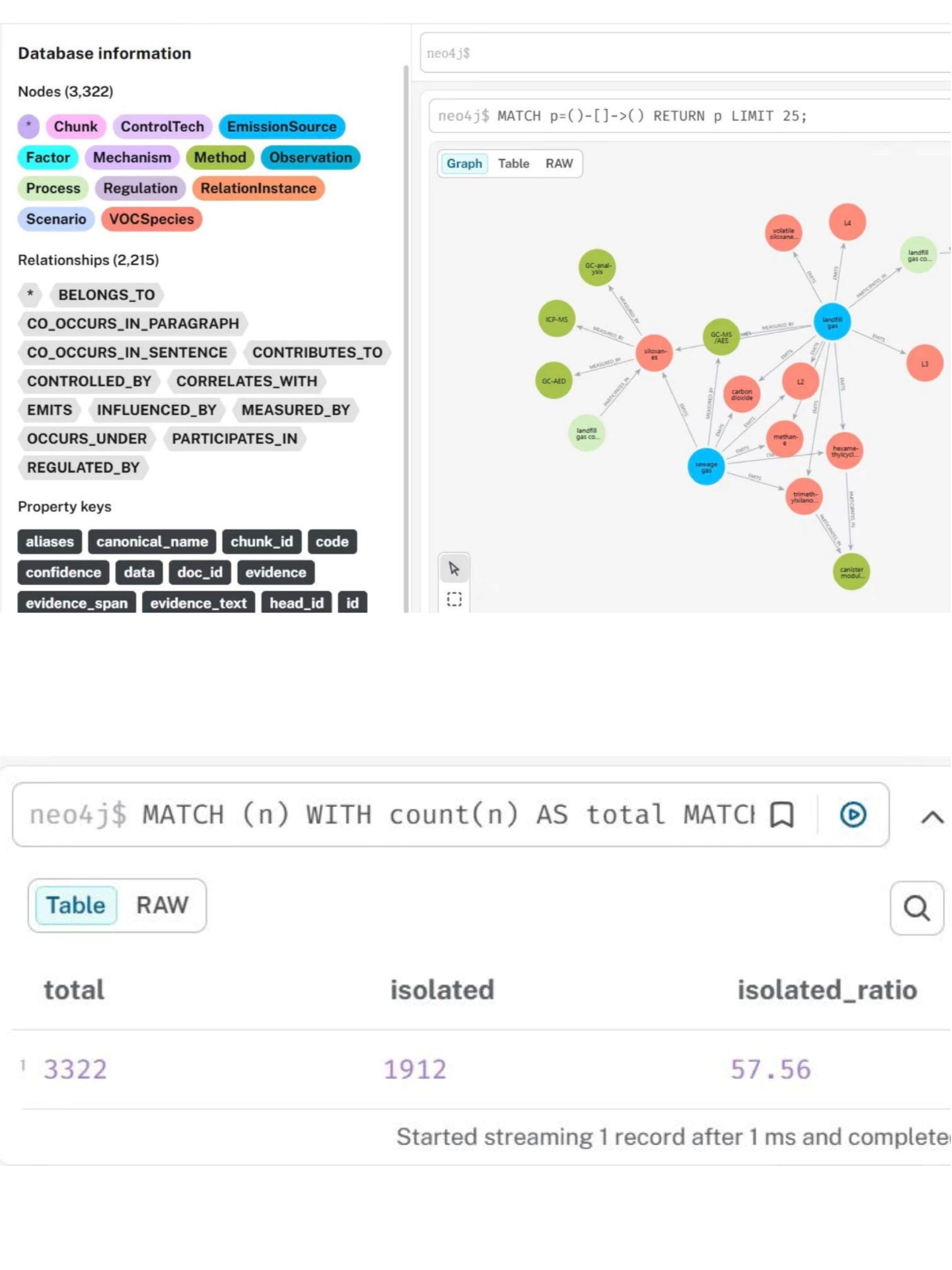}
  \caption{Snapshot of the initial knowledge-graph topology generated by the baseline LLM extraction pipeline. The visualization shows severe fragmentation with an isolated node rate of approximately 57 percent.}
  \label{fig:s_part}
\end{figure}

\clearpage
\addcontentsline{toc}{subsection}{Figure S4. Expert Evaluation Scores and Consistency Analysis}
\begin{figure}[H]
  \centering
  \includegraphics[width=1\textwidth,height=0.78\textheight,keepaspectratio]{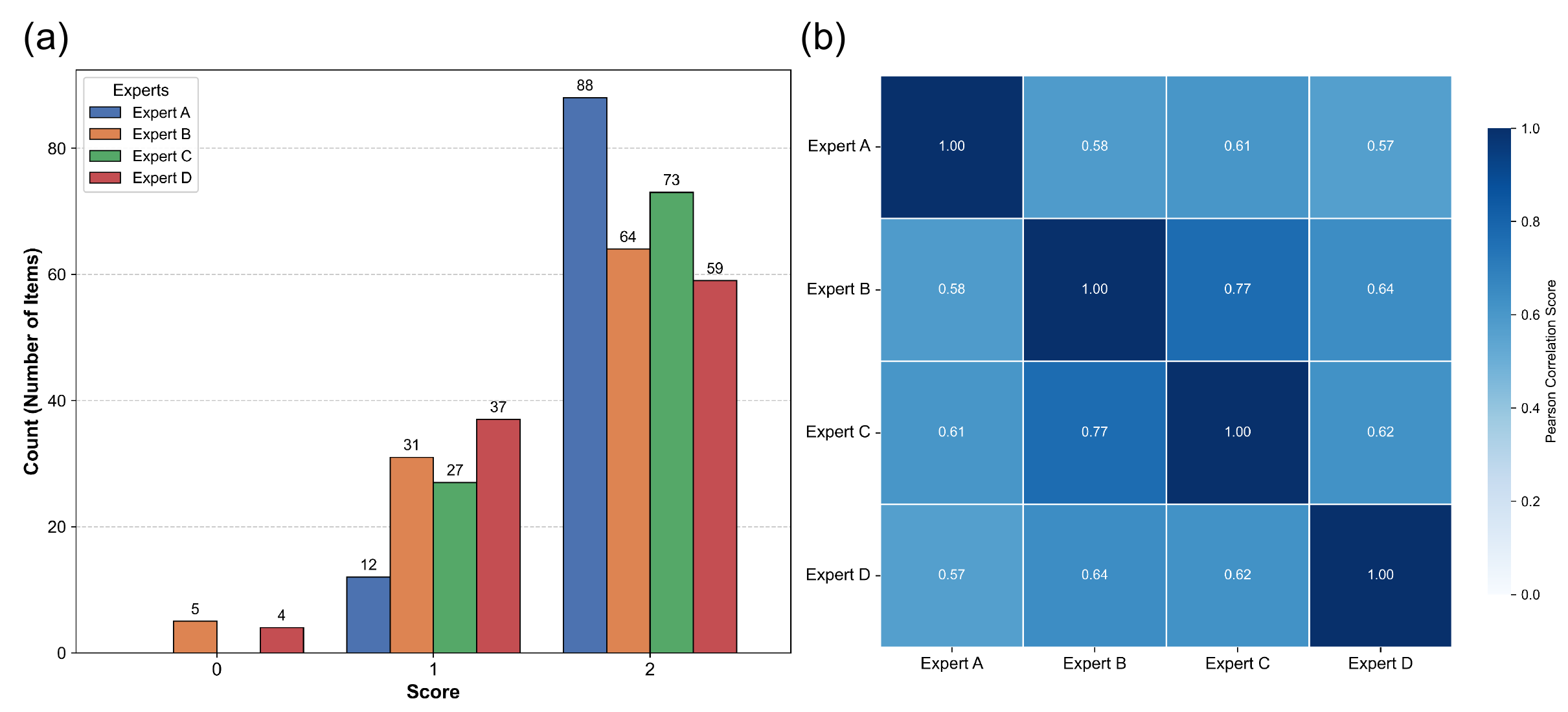}
  \caption{Comprehensive quantitative evaluation and scoring consistency analysis based on expert blind assessment. Distribution of expert evaluation scores detailing the count of items assigned 0, 1, and 2 points by each of the four experts (a); Expert scoring consensus analysis presenting the Pearson correlation matrix across the four experts (b).}
  \label{fig:s_perf1}
\end{figure}

\clearpage
\addcontentsline{toc}{subsection}{Figure S5. Generalization-Performance Evaluation}
\begin{figure}[H]
  \centering
  \includegraphics[width=1\textwidth,height=0.78\textheight,keepaspectratio]{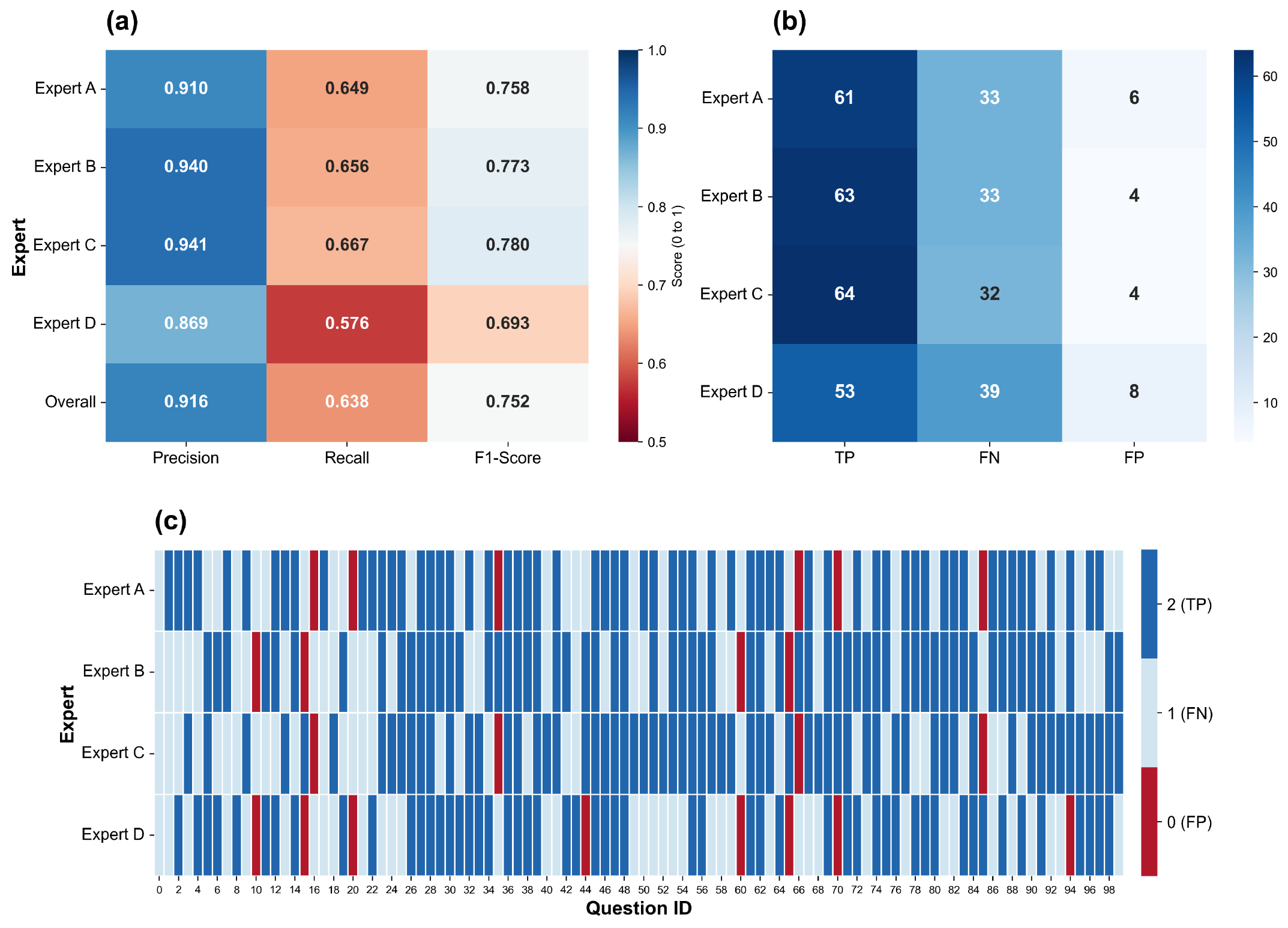}
  \caption{Generalization performance evaluation of the multi-agent system across broad environmental engineering topics and cross domain industrial scenarios. Heatmap of Precision, Recall, and F1-score evaluated independently by four experts (a); Count distribution matrix of true positives, false negatives, and false positives (b); Panoramic expert consensus matrix recording individual scoring variations across 100 extended domain questions (c).}
  \label{fig:s_perf2}
\end{figure}

\clearpage
\addcontentsline{toc}{subsection}{Table S1. Literature Search Strategy and Data Distribution}
\begin{table}[H]
  \caption{Literature Search Strategy and Data Distribution in the Steel-Industry VOC Domain.}
  \label{tbl:s1_search_strategy}
  \begin{tabular}{p{0.30\linewidth}p{0.48\linewidth}p{0.16\linewidth}}
    \hline
    Search Dimension & Search Logic & Initially Retrieved Papers \\
    \hline
    Macro emissions from steel industry & TS = ("Steel industry" AND "VOCs emissions") & 63 \\
    Emission sources and composition spectrum & TS = ("Steel industry" AND "Source composition spectrum") & 67 \\
    Emissions from ironmaking process & TS = ("ironmaking" AND "VOCs emissions") & 1 \\
    Emissions from sintering process & TS = ("sintering" AND "VOCs emissions") & 46 \\
    Emissions from coking process & TS = ("coking" AND "VOCs emissions") & 169 \\
    Total & -- & 346 \\
    Final corpus size & -- & 382 \\
    \hline
  \end{tabular}
\end{table}

\clearpage
\addcontentsline{toc}{subsection}{Table S2. Ontology Schema and Representative Definitions}
\begin{table}[H]
  \caption{Ontology schema and representative definitions of entity categories and relation types used in the steel-industry VOC knowledge graph.}
  \label{tbl:s2_ontology}
  \begin{tabular}{p{0.14\linewidth}p{0.20\linewidth}p{0.60\linewidth}}
    \hline
    Category & Label & Technical Definition and Example \\
    \hline
    Entity & Process & Core production stages (e.g., sintering, coking, cold rolling) \\
    Entity & EmissionSource & Specific equipment or release points (e.g., main stack, coke oven battery) \\
    Entity & VOCSpecies & Specific compounds or classes (e.g., benzene, alkanes, PAHs) \\
    Entity & ControlTech & Abatement technologies (e.g., regenerative thermal oxidation, activated carbon adsorption) \\
    Entity & Method & Sampling and analytical protocols (e.g., GC--MS, FID, PTR-TOF-MS) \\
    Entity & Mechanism & Chemical or physical mechanisms (e.g., catalytic degradation, photo-oxidation) \\
    Entity & Regulation & Emission standards and policy constraints (e.g., GB~28662-2012) \\
    Entity & Factor & Influencing parameters (e.g., operating temperature, humidity, space velocity) \\
    Entity & Scenario & Specific operating contexts (e.g., start-up phase, abnormal conditions) \\
    Relation & emits & Direct link between a process/source and pollutant species \\
    Relation & controlled\_by & Link between a process/source and abatement technology \\
    Relation & measured\_by & Link between a pollutant/source and analytical method \\
    Relation & participates\_in & Participation of a species/source in a mechanism or process \\
    Relation & influenced\_by & Influence of factors/scenarios on emissions or control efficiency \\
    Relation & belongs\_to & Taxonomic hierarchy (e.g., toluene belongs to aromatics) \\
    Relation & co\_occurs\_in & Heuristic association based on sentence/paragraph co-occurrence (anti-isolation rule) \\
    \hline
  \end{tabular}
\end{table}

\clearpage
\addcontentsline{toc}{subsection}{Table S3. Detailed Node Distribution}
\begin{table}[H]
  \caption{Detailed node distribution of the steel-industry VOC knowledge graph.}
  \label{tbl:s2_node_distribution}
  \begin{tabular}{p{0.16\linewidth}p{0.18\linewidth}p{0.14\linewidth}p{0.14\linewidth}p{0.30\linewidth}}
    \hline
    Label & Semantic Description & Stage~1 (49.88\%) & Stage~2 (4.08\%) & Optimization Effect and Notes \\
    \hline
    Observation$^{*}$ & Observational data points & 45,662 & 1,103 & Long-tail numeric nodes were extensively collapsed through entity normalization and attribute conversion. \\
    EmissionSource & Emission sources/equipment & 5,812 & 3,936 & Alias mapping removed duplicate names for identical sources. \\
    VOCSpecies & VOC species/classes & 4,812 & 3,781 & Alias mapping removed duplicate naming variants. \\
    Process & Production processes & 1,583 & 3,656 & Finer chunk-level restructuring improved process-level entity resolution. \\
    Method & Monitoring & 4,285 & 3,005 & Normalization reduced redundancy. \\
    Factor & Influencing factors & 5,122 & 1,367 & Normalization reduced noisy long-tail variants. \\
    ControlTech & Control technologies & 1,809 & 640 & Core technologies converged to standardized terminology. \\
    Regulation & Regulatory standards & 1,074 & 599 & Standardization and de-duplication. \\
    Scenario & Operating scenarios & 1,817 & 589 & Standardization and de-duplication. \\
    Mechanism & Reaction mechanisms & 991 & 113 & High-level abstraction and consolidation. \\
    Chunk & Literature chunks & 0 & 8,391 & Newly introduced hub node type in Stage~2. \\
    Total & -- & 72,967 & 27,180 & Global topology achieved high-dimensional denoising and dense restructuring. \\
    \hline
  \end{tabular}
\end{table}

\clearpage
\addcontentsline{toc}{subsection}{Table S4. Detailed Relationship Distribution}
\begin{table}[H]
  \caption{Detailed relationship distribution of the knowledge graph.}
  \label{tbl:s3_relation_distribution}
  \scriptsize
  \setlength{\tabcolsep}{3pt}
  \begin{tabular}{>{\raggedright\arraybackslash}p{0.22\linewidth}>{\raggedright\arraybackslash}p{0.18\linewidth}>{\raggedright\arraybackslash}p{0.12\linewidth}>{\raggedright\arraybackslash}p{0.12\linewidth}>{\raggedright\arraybackslash}p{0.28\linewidth}}
    \hline
    Relation Type & Semantic Meaning & Stage~1 (49.88\%) & Stage~2 (4.08\%) & Evolution Note \\
    \hline
    EMITS & Emission linkage & 10,194 & 10,437 & Core emission skeleton preserved and slightly expanded. \\
    PARTICIPATES\_IN & Mechanism participation & 1,720 & 4,669 & Improved discovery of deeper mechanistic pathways. \\
    INFLUENCED\_BY & Factor influence & 4,566 & 4,108 & Mild reduction after entity normalization. \\
    CONTROLLED\_BY$^{*}$ & Control linkage & 1,440 & 2,582 & Increased after model-assisted inference was incorporated. \\
    MEASURED\_BY & Measurement linkage & 3,002 & 1,978 & Reconstructed after compressing redundant numeric nodes. \\
    CORRELATES\_WITH & Correlative linkage & 2,043 & 580 & Weak semantic edges replaced by higher-confidence links. \\
    REGULATED\_BY & Regulatory constraint & 953 & 394 & Reduced with regulation-entity consolidation. \\
    OCCURS\_UNDER & Scenario constraint & 905 & 309 & Reduced with scenario-entity consolidation. \\
    BELONGS\_TO & Taxonomic linkage & 1,229 & 98 & Flattening removed over-nested hierarchy segments. \\
    MEASURES & Early numeric edge type & 20,979 & 0 & Fully absorbed by schema restructuring and attribute migration. \\
    Other edges & Long-tail/co-occurrence edges & 229 & 2,369 & Added anti-isolation long-tail rules (e.g., paragraph-level co-occurrence). \\
    :MENTIONS & Traceable chunk mention linkage & 0 & 54,255 & Critical back-end linkage between entities and source chunks. \\
    Total & -- & 47,260 & 81,779 & Substantial increase in traceable and reasoning-ready connectivity. \\
    \hline
  \end{tabular}
\end{table}

\clearpage
\addcontentsline{toc}{subsection}{Table S5. Multi-Model Comparison}
\begin{table}[H]
  \caption{Multi-model comparison on complex industrial pollutant reasoning.}
  \label{tbl:s4_model_comparison}
  \scriptsize
  \setlength{\tabcolsep}{4pt}
  \begin{tabular}{>{\raggedright\arraybackslash}p{0.16\linewidth}>{\raggedright\arraybackslash}p{0.78\linewidth}}
    \hline
    Model & Response excerpt \\
    \hline
    User question & What are the key characteristic VOC species emitted from the sintering process within the steel industry? \\
    Chat-ISV & The key characteristic volatile organic compounds (VOCs) emitted from the sintering process within the steel industry include several major families and representative species. Based on the retrieved evidence, the response highlighted alkanes, alkenes, aromatics, halohydrocarbons, chloromethane, chloroethane, and non-methane hydrocarbons (NMHC), together with quantitative evidence such as an NMHC emission factor of 11.14 g/t and a chloromethane emission factor of 0.9973 g/t. The answer also noted that halohydrocarbons account for about 25\% of emissions and discussed the broader mixed-emission profile of the sintering process. \\
    GPT-4o-mini & The response stated that VOC emissions from the sintering process are characterized primarily by aromatic hydrocarbons and polycyclic aromatic hydrocarbons (PAHs), particularly BTEX compounds, together with OVOCs and light alkenes. It further emphasized benzene, toluene, ethylbenzene, xylene, naphthalene, phenanthrene, benzo[a]pyrene, formaldehyde, acetaldehyde, ethylene, and propylene as characteristic species. \\
    GPT-4o & The response described sintering emissions as a complex mixture dominated by aromatic hydrocarbons, PAHs, OVOCs, aliphatic hydrocarbons, and in some cases trace halogenated hydrocarbons. Representative species listed included BTEX compounds, naphthalene, phenanthrene, pyrene, formaldehyde, acetaldehyde, acetone, ethylene, and propylene, and the model concluded that BTEX and PAHs are the core characteristic species. \\
    Claude-3.5-Sonnet & The response organized the answer into PAHs, BTEX compounds, aliphatic hydrocarbons, oxygenated VOCs, chlorinated compounds, and heterocyclic compounds. It emphasized naphthalene, fluorene, phenanthrene, anthracene, benzene, toluene, xylenes, formaldehyde, acetaldehyde, chlorobenzenes, chlorophenols, furans, pyridine, and quinoline, while also discussing operating factors and regulatory relevance. \\
    Kimi-K2.5 & The response identified aliphatic hydrocarbons, BTEX and PAHs, oxygenated VOCs, halogenated compounds, sulfur-containing VOCs, and nitrogen-containing VOCs. It listed examples such as methane, ethane, propane, ethylene, propylene, benzene, toluene, xylenes, naphthalene, formaldehyde, acetaldehyde, chloromethane, dichloromethane, thiophene, pyridine, and acetonitrile, and related the profile to feedstock and operating conditions. \\
    DeepSeek-chat & The response stated that sintering emissions are predominantly aromatic hydrocarbons, especially BTEX, and further highlighted carbon disulfide as a distinctive marker. Additional examples included isopentane, 1,2,4-trimethylbenzene, and methyl bromide, with the answer framing these species as major contributors to ozone formation and VOC-control priorities. \\
    Qwen-Max & The response summarized characteristic species as alkanes, alkenes, benzene-series compounds, halogenated hydrocarbons, PAHs, and OVOCs. It attributed these species to incomplete combustion and pyrolysis of solid fuels and identified benzene-series compounds as major control targets while emphasizing the atmospheric reactivity of alkanes and alkenes. \\
    Gemini-pro & The response argued that oxygenated VOCs and aromatic hydrocarbons dominate steel-sintering flue gas, and highlighted acetone, formaldehyde, benzene, toluene, ethylbenzene, short-chain alkanes, alkenes, and chloromethane as representative species. It further described these compounds as key contributors to odor, ozone, and PM$_{2.5}$ formation. \\
    GLM-4 & The response provided a long categorized answer covering chlorinated VOCs, oxygenated VOCs, aromatic hydrocarbons, and light aliphatic hydrocarbons. Representative species included chlorobenzene, dichloromethane, chloroform, vinyl chloride, phenol, cresols, acetone, benzaldehyde, formaldehyde, benzene, toluene, xylenes, methane, ethane, and propene, with additional discussion of feedstock composition, oxygen availability, residence time, and environmental impacts. \\
    \hline
  \end{tabular}
  \normalsize
\end{table}

\clearpage
\addcontentsline{toc}{subsection}{Table S6. Raw Expert Blind-Evaluation Dataset Sample}
\begin{table}[H]
  \caption{Sample of the raw expert blind-evaluation dataset.}
  \label{tbl:s5_expert_sample}
  \begin{tabular}{p{0.06\linewidth}p{0.23\linewidth}p{0.44\linewidth}p{0.06\linewidth}p{0.06\linewidth}p{0.06\linewidth}p{0.06\linewidth}}
    \hline
    ID & Test Question & System Output / Context (Excerpt) & Exp. A & Exp. B & Exp. C & Exp. D \\
    \hline
    1 & What is the average NMHC emission factor for the steel-industry sintering process? & ``...the context indicates NMHC emissions from sintering... but does not provide a complete direct emission-factor statement...'' & 1 & 0 & 1 & 0 \\
    100 & Does the sintering process produce VOCs? & ``Yes, the iron-ore sintering process produces VOCs...'' & 2 & 2 & 2 & 2 \\
    \hline
  \end{tabular}
\end{table}

\nolinenumbers
\clearpage
\bibliographystyle{abbrv}
\bibliography{library}

@ARTICLE{Zhang2022SynergisticControl,
  author = {Zhang, Y. and others},
  title = {Emission Characteristics and Synergistic Control Technologies of Volatile Organic Compounds in Typical Industrial Processes},
  journal = {Journal of Hazardous Materials},
  year = {2022},
  volume = {424},
  pages = {127645},
  doi = {10.1016/j.jhazmat.2021.127645},
}

@ARTICLE{Wu2025PromptHallucination,
  author = {Wu, X. and others},
  title = {Augmented and Programmatically Optimized LLM Prompts Reduce Chemical Hallucinations},
  journal = {Journal of Chemical Information and Modeling},
  year = {2025},
  note = {Published online ahead of print},
  doi = {10.1021/acs.jcim.4c02322},
}

@ARTICLE{Zhou2024CausalKGPT,
  author = {Zhou, B. and Li, X. and Liu, T. and Xu, K. and Liu, W. and Bao, J.},
  title = {CausalKGPT: Industrial Structure Causal Knowledge-Enhanced Large Language Model for Cause Analysis of Quality Problems in Aerospace Product Manufacturing},
  journal = {Advanced Engineering Informatics},
  year = {2024},
  volume = {59},
  pages = {102333},
  doi = {10.1016/j.aei.2023.102333},
}

@ARTICLE{Tang2022EnvRes,
  author = {Tang, L. and others},
  title = {Characteristics and Health Risks of Volatile Organic Compounds Emitted from a Typical Iron and Steel Industry in North China},
  journal = {Environ. Res.},
  year = {2022},
  volume = {212},
  pages = {113337},
  doi = {10.1016/j.envres.2022.113337},
}

@ARTICLE{Chen2024STOTEN,
  author = {Chen, Z. and others},
  title = {Multi-Pollutant Collaborative Control Technologies in the Iron and Steel Sintering Process: Status and Perspectives},
  journal = {Sci. Total Environ.},
  year = {2024},
  volume = {908},
  pages = {168291},
  doi = {10.1016/j.scitotenv.2023.168291},
}

@ARTICLE{Zhang2016ChemRevVOC,
  author = {Zhang, X. and others},
  title = {Recent Advances in the Catalytic Oxidation of Volatile Organic Compounds: A Review Based on Pollutant Sorts and Sources},
  journal = {Chemical Reviews},
  year = {2016},
  volume = {116},
  number = {6},
  pages = {3622--3673},
  doi = {10.1021/acs.chemrev.5b00416},
}

@ARTICLE{Zheng2023ChatGPTChem,
  author = {Zheng, Z. and others},
  title = {ChatGPT in Chemical Research and Education},
  journal = {ACS Central Science},
  year = {2023},
  volume = {9},
  number = {8},
  pages = {1459--1466},
  doi = {10.1021/acscentsci.3c00446},
}

@ARTICLE{Bran2024ChemCrow,
  author = {Bran, A. M. and others},
  title = {ChemCrow: Augmenting Large-Language Models with Chemistry Tools},
  journal = {Nature Machine Intelligence},
  year = {2024},
  volume = {6},
  number = {5},
  pages = {481--487},
  doi = {10.1038/s42256-024-00832-8},
}

@ARTICLE{Ji2023HallucinationSurvey,
  author = {Ji, Z. and others},
  title = {Survey of Hallucination in Natural Language Generation},
  journal = {ACM Computing Surveys},
  year = {2023},
  volume = {55},
  number = {12},
  pages = {1--38},
  doi = {10.1145/3571730},
}

@ARTICLE{White2023FutureChem,
  author = {White, A. D. and others},
  title = {The Future of Chemistry Is Language},
  journal = {Nature Reviews Chemistry},
  year = {2023},
  volume = {7},
  number = {7},
  pages = {457--458},
  doi = {10.1038/s41570-023-00502-0},
}

@ARTICLE{Bai2025npjCompMat,
  author = {Bai, X. and He, S. and Li, Y. and Xie, Y. and Zhang, X. and Du, W. and Li, J.-R.},
  title = {Construction of a Knowledge Graph for Framework Material Enabled by Large Language Models and Its Application},
  journal = {npj Computational Materials},
  year = {2025},
  volume = {11},
  pages = {51},
  doi = {10.1038/s41524-025-01540-6},
}

@ARTICLE{Dreger2025DigitDiscov,
  author = {Dreger, M. and Malek, K. and Eikerling, M.},
  title = {Large Language Models for Knowledge Graph Extraction from Tables in Materials Science},
  journal = {Digital Discovery},
  year = {2025},
  volume = {4},
  pages = {1221},
  doi = {10.1039/D4DD00362D},
}

@ARTICLE{Miret2025CSR,
  author = {Miret, S. and others},
  title = {From Text to Insight: Large Language Models for Chemical Data Extraction},
  journal = {Chemical Society Reviews},
  year = {2025},
  volume = {54},
  number = {3},
  pages = {1125--1150},
  doi = {10.1039/D4CS00913D},
}

@ARTICLE{Pan2024TKDE,
  author = {Pan, Shirui and Luo, Linhao and Wang, Yufei and Chen, Chen and Wang, Jiapu and Wu, Xindong},
  title = {Unifying Large Language Models and Knowledge Graphs: A Roadmap},
  journal = {IEEE Transactions on Knowledge and Data Engineering},
  year = {2024},
  volume = {36},
  number = {8},
  pages = {3534--3554},
  doi = {10.1109/TKDE.2024.3352100},
}

@ARTICLE{Lewis2020RAG,
  author = {Lewis, P. and others},
  title = {Retrieval-Augmented Generation for Knowledge-Intensive {NLP} Tasks},
  journal = {Advances in Neural Information Processing Systems},
  year = {2020},
  volume = {33},
  pages = {9459--9474},
  eprint = {2005.11401},
  archivePrefix = {arXiv},
}

@ARTICLE{Hogan2021KG,
  author = {Hogan, A. and others},
  title = {Knowledge Graphs},
  journal = {ACM Computing Surveys},
  year = {2021},
  volume = {54},
  number = {4},
  pages = {1--37},
  doi = {10.1145/3447772},
}

@ARTICLE{Wang2026ChatRFB,
  author = {Wang, H.-T. and Bai, X. and Zheng, Z. and Zhang, X. and Jin, R. and An, H.-T. and Xie, Z.-H. and Lv, X.-L. and Li, J.-R.},
  title = {Chat-RFB: A Flow Battery Chat System Leveraging Knowledge Graphs and Large Language Models},
  journal = {Digital Discovery},
  year = {2026},
  volume = {5},
  pages = {1401--1410},
  doi = {10.1039/d5dd00494b},
}

@ARTICLE{Wang2024AgentSurvey,
  author = {Wang, L. and others},
  title = {A Survey on Large Language Model Based Autonomous Agents},
  journal = {Frontiers of Computer Science},
  year = {2024},
  volume = {18},
  number = {6},
  pages = {186345},
  doi = {10.1007/s11704-024-40231-1},
}

@ARTICLE{Ouyang2022InstructGPT,
  author = {Ouyang, L. and others},
  title = {Training Language Models to Follow Instructions with Human Feedback},
  journal = {Advances in Neural Information Processing Systems},
  year = {2022},
  volume = {35},
  pages = {27730--27744},
  eprint = {2203.02155},
  archivePrefix = {arXiv},
}

@ARTICLE{Edge2024GraphRAG,
  author = {Edge, D. and others},
  title = {From Local to Global: A Graph {RAG} Approach to Query-Focused Summarization},
  journal = {arXiv},
  year = {2024},
  eprint = {2404.16130},
  archivePrefix = {arXiv},
}

@ARTICLE{Gao2023RAGSurvey,
  author = {Gao, Y. and others},
  title = {Retrieval-Augmented Generation for Large Language Models: A Survey},
  journal = {arXiv},
  year = {2023},
  eprint = {2312.10997},
  archivePrefix = {arXiv},
}

@ARTICLE{Wu2023AutoGen,
  author = {Wu, Q. and others},
  title = {AutoGen: Enabling Next-Gen {LLM} Applications via Multi-Agent Conversation},
  journal = {arXiv},
  year = {2023},
  eprint = {2308.08155},
  archivePrefix = {arXiv},
}

@ARTICLE{Agrawal2024Patterns,
  author = {Agrawal, A. and others},
  title = {Large Language Models for Knowledge Graph Extraction from Scientific Literature},
  journal = {Patterns},
  year = {2024},
  volume = {5},
  number = {1},
  pages = {100913},
}

@INPROCEEDINGS{Yao2023ReAct,
  author = {Yao, S. and others},
  title = {ReAct: Synergizing Reasoning and Acting in Language Models},
  booktitle = {International Conference on Learning Representations},
  year = {2023},
  eprint = {2210.03629},
  archivePrefix = {arXiv},
}

@ARTICLE{Huang2023HallucinationSurvey2,
  author = {Huang, L. and others},
  title = {A Survey on Hallucination in Large Language Models: Principles, Taxonomy, Challenges, and Open Questions},
  journal = {arXiv},
  year = {2023},
  eprint = {2311.05232},
  archivePrefix = {arXiv},
}

@ARTICLE{Taylor2022Galactica,
  author = {Taylor, R. and others},
  title = {Galactica: A Large Language Model for Science},
  journal = {arXiv},
  year = {2022},
  eprint = {2211.09085},
  archivePrefix = {arXiv},
}

@ARTICLE{Zhu2024LLMsfor,
  author = {Zhu, Yuqi and Wang, Xiaohan and Chen, Jing and Qiao, Shuofei and Ou, Yixin and Yao, Yunzhi and Deng, Shumin and Chen, Huajun and Zhang, Ningyu},
  title = {LLMs for knowledge graph construction and reasoning: recent capabilities and future opportunities},
  journal = {WORLD WIDE WEB-INTERNET AND WEB INFORMATION SYSTEMS},
  year = {2024},
  volume = {27},
  number = {5},
  pages = {58},
  doi = {10.1007/s11280-024-01297-w},
}

@ARTICLE{Xu2024ChatTfA,
  author = {Xu, Jun and Zhang, Hao and Zhang, Haijing and Lu, Jiawei and Xiao, Gang},
  title = {ChatTf: A Knowledge Graph-Enhanced Intelligent Q\&A System for Mitigating Factuality Hallucinations in Traditional Folklore},
  journal = {IEEE ACCESS},
  year = {2024},
  volume = {12},
  pages = {162638-162650},
  doi = {10.1109/ACCESS.2024.3485877},
}

@ARTICLE{Duan2025LLMempowered,
  author = {Duan, Yifei and Tian, Yixi and Ghosh, Soumya and Venugopal, Vineeth and Chen, Jie and Olivetti, Elsa A.},
  title = {LLM-empowered literature mining for material substitution studies in sustainable concrete},
  journal = {RESOURCES CONSERVATION AND RECYCLING},
  year = {2025},
  volume = {221},
  pages = {108379},
  doi = {10.1016/j.resconrec.2025.108379},
}

@ARTICLE{Kumar2025ALarge,
  author = {Kumar, Avan and Nazemi, Farshid and Kodamana, Hariprasad and Ramteke, Manojkumar and Bakshi, Bhavik R.},
  title = {A Large Language Model-based Framework to Retrieve Life Cycle Inventory and Environmental Impact Data from Scientific Literature},
  journal = {ENVIRONMENTAL SCIENCE \& TECHNOLOGY},
  year = {2025},
  volume = {59},
  number = {42},
  pages = {22533-22543},
  doi = {10.1021/acs.est.5c05955},
}

@ARTICLE{Wang2025LLMKGMQA,
  author = {Wang, FeiLong and Shi, Donghui and Aguilar, Jose and Cui, Xinyi and Jiang, Jinsong and Shen, Longjian and Li, Mengya},
  title = {LLM-KGMQA: large language model-augmented multi-hop question-answering system based on knowledge graph in medical field},
  journal = {KNOWLEDGE AND INFORMATION SYSTEMS},
  year = {2025},
  volume = {67},
  number = {8},
  pages = {6461-6503},
  doi = {10.1007/s10115-025-02399-1},
}

@ARTICLE{He2024GRetriever,
  author = {He, Xiaoxin and Tian, Yijun and Sun, Yifei and Chawla, Nitesh V. and Laurent, Thomas and LeCun, Yann and Bresson, Xavier and Hooi, Bryan},
  title = {G-Retriever: Retrieval-Augmented Generation for Textual Graph Understanding and Question Answering},
  journal = {Advances in Neural Information Processing Systems},
  year = {2024},
  volume = {37},
  eprint = {2402.07630},
  archivePrefix = {arXiv},
  doi = {10.48550/arXiv.2402.07630},
}

@ARTICLE{Sun2024ToG,
  author = {Sun, Jiashuo and Xu, Chengjin and Tang, Lumingyuan and Wang, Saizhuo and Lin, Chen and Gong, Yeyun and Ni, Lionel M. and Shum, Heung-Yeung and Guo, Jian},
  title = {Think-on-Graph: Deep and Responsible Reasoning of Large Language Model on Knowledge Graph},
  journal = {International Conference on Learning Representations},
  year = {2024},
  eprint = {2307.07697},
  archivePrefix = {arXiv},
  doi = {10.48550/arXiv.2307.07697},
}

@ARTICLE{Zhou2025MethaneFeAl,
  author = {Li, Shuang and Liao, Junyi and Zhang, Zhanguo and Xu, Guangwen},
  title = {Catalytic decomposition of methane over {Fe2O3-Al2O3} catalysts with high iron contents and at high {CH4} space velocities},
  journal = {Resources Chemicals and Materials},
  year = {2025},
  volume = {4},
  number = {4},
  pages = {100123},
  doi = {10.1016/j.recm.2025.100123},
}

@ARTICLE{Lin2026MOFMultiAgent,
  author = {Lin, Zuhong and Ren, Daoyuan and Ran, Kai and Sun, Jing and Yu, Songlin and Bai, Xuefeng and Huang, Xiaotian and He, Haiyang and Pan, Pengxu and Fang, Ying and Li, Zhanglin and Li, Haipu and Yao, Jingjing},
  title = {Reshaping {MOFs} synthesis conditions mining with a dynamic multi-agents framework of large language model},
  journal = {Transactions of Materials Research},
  year = {2026},
  volume = {2},
  number = {1},
  pages = {100176},
  doi = {10.1016/j.tramat.2026.100176},
}

@ARTICLE{Zheng2026ExpertSystems,
  author = {Zheng, Peng and Xu, Guangwen},
  title = {Expert systems: Grounding cross-disciplinary {LLMs} in reality},
  journal = {Resources Chemicals and Materials},
  year = {2026},
  volume = {5},
  number = {1},
  pages = {100164},
  doi = {10.1016/j.recm.2025.100164},
}

@ARTICLE{Lin2025MOFArsenateGCN,
  author = {Lin, Zuhong and Chen, Jiarong and Fang, Ying and Deng, Shi-hai and Li, Haipu and Yang, Ying and Yao, Jingjing},
  title = {Rapidly tailor metal--organic frameworks for arsenate removal using graph convolutional neural networks},
  journal = {Separation and Purification Technology},
  year = {2025},
  volume = {354},
  pages = {129334},
  doi = {10.1016/j.seppur.2024.129334},
}

@MISC{Ding2026ChatISVDataset,
  author = {Ding, Yu},
  title = {Dataset for {Chat-ISV}: A Knowledge Graph-Enhanced Large Language Model for Question Answering on {VOC} Emission Control in the Steel Industry},
  howpublished = {Zenodo},
  year = {2026},
  doi = {10.5281/zenodo.20280089},
}

@ARTICLE{Ji2022KGSurvey,
  author = {Ji, Shaoxiong and Pan, Shirui and Cambria, Erik and Marttinen, Pekka and Yu, Philip S.},
  title = {A Survey on Knowledge Graphs: Representation, Acquisition, and Applications},
  journal = {IEEE Transactions on Neural Networks and Learning Systems},
  year = {2022},
  volume = {33},
  number = {2},
  pages = {494--514},
  doi = {10.1109/TNNLS.2021.3070843},
}

@ARTICLE{Talebirad2023MultiAgent,
  author = {Talebirad, Yashar and Nadiri, Amirhossein},
  title = {Multi-Agent Collaboration: Harnessing the Power of Intelligent {LLM} Agents},
  journal = {arXiv preprint arXiv:2306.03314},
  year = {2023},
  eprint = {2306.03314},
  archivePrefix = {arXiv},
}

@INPROCEEDINGS{Lan2024AgentSociety,
  author = {Lan, Yihuai and Hu, Zhiqiang and Wang, Lei and Wang, Yang and Ye, Deheng and Zhao, Peilin and Lim, Ee-Peng and Xiong, Hui and Wang, Hao},
  title = {{LLM}-Based Agent Society Investigation: Collaboration and Confrontation in {Avalon} Gameplay},
  booktitle = {Proceedings of the 2024 Conference on Empirical Methods in Natural Language Processing},
  year = {2024},
  pages = {100--113},
  doi = {10.18653/v1/2024.emnlp-main.7},
}

@INPROCEEDINGS{Wang2024KGPrompting,
  author = {Wang, Yu and Lipka, Nedim and Rossi, Ryan A. and Siu, Alexa and Zhang, Ruiyi and Derr, Tyler},
  title = {Knowledge Graph Prompting for Multi-Document Question Answering},
  booktitle = {Proceedings of the AAAI Conference on Artificial Intelligence},
  year = {2024},
  volume = {38},
  number = {17},
  pages = {19206--19214},
  doi = {10.1609/aaai.v38i17.29889},
}

@ARTICLE{Zhang2019CoalPolicy,
  author = {Zhang, Yixuan and Bo, Xian and Zhao, Yu and Nielsen, Chris P. and Zhang, Jing and Zhang, Shaohui and Li, Xue and Zhao, Bao and Geng, Guannan and Zhang, Qiang},
  title = {Benefits of current and future policies on emissions of China's coal-fired power sector indicated by continuous emission monitoring},
  journal = {Environmental Pollution},
  year = {2019},
  volume = {251},
  pages = {415--424},
  doi = {10.1016/j.envpol.2019.05.021},
}

@ARTICLE{Su2025CJCHE,
  author = {Su, Changqing and Jiang, Wentao and Guo, Yang and Yi, Guodong and Li, Zengxing and Li, Huan},
  title = {Rational molecular design of {P}-doped porous carbon material for the {VOCs} adsorption},
  journal = {Chinese Journal of Chemical Engineering},
  year = {2025},
  volume = {79},
  pages = {155--163},
  doi = {10.1016/j.cjche.2024.11.017},
}

@ARTICLE{Su2020MaterChemPhys,
  author = {Su, Changqing and Guo, Yang and Yu, Lingyun and Zou, Jianwu and Zeng, Zheng and Li, Liqing},
  title = {Insight into specific surface area, microporosity and {N}, {P} co-doping of porous carbon materials in the acetone adsorption},
  journal = {Materials Chemistry and Physics},
  year = {2020},
  volume = {258},
  pages = {123930},
  doi = {10.1016/j.matchemphys.2020.123930},
}

@ARTICLE{Su2020Colloids,
  author = {Su, Changqing and Guo, Yang and Chen, Hongyu and Zou, Jianwu and Zeng, Zheng and Li, Liqing},
  title = {{VOCs} adsorption of resin-based activated carbon and bamboo char: porous characterization and nitrogen-doped effect},
  journal = {Colloids and Surfaces A: Physicochemical and Engineering Aspects},
  year = {2020},
  volume = {601},
  pages = {124983},
  doi = {10.1016/j.colsurfa.2020.124983},
}

@ARTICLE{Jafarzadeh2024AKG,
  author = {Jafarzadeh, Parastoo and Ensan, Faezeh and Alavi, Mahdiyar Ali Akbar and Zarrinkalam, Fattane},
  title = {A Knowledge Graph Embedding Model for Answering Factoid Entity Questions},
  journal = {ACM Transactions on Information Systems},
  year = {2024},
  volume = {43},
  number = {2},
  pages = {Article 34},
  doi = {10.1145/3678003},
}

@ARTICLE{Li2025TKDE,
  author = {Li, Yading and Song, Dandan and Tian, Yuhang and Wang, Hao and Zhou, Changzhi and Zhang, Shuhao},
  title = {A Framework of Knowledge Graph-Enhanced Large Language Model Based on Global Planning},
  journal = {IEEE Transactions on Knowledge and Data Engineering},
  year = {2026},
  volume = {38},
  number = {2},
  pages = {736--748},
  doi = {10.1109/TKDE.2025.3639599},
}

@ARTICLE{Lin2026LowCarbonScheduling,
  author = {Lin, M. and Wu, T. and Xie, K.},
  title = {Low-Carbon Scheduling for Power-Hydrogen Integrated Energy System Using Large Language Model Enhanced Deep Reinforcement Learning},
  journal = {IEEE Transactions on Sustainable Energy},
  year = {2026},
  doi = {10.1109/TSTE.2026.3677475},
}

@ARTICLE{Liu2022JEM,
  author = {Liu, Zewei and others},
  title = {Characterizing the Emission Behaviors of Cumulative {VOCs} from Automotive Solvent-Based Paint Sludge},
  journal = {Journal of Environmental Management},
  year = {2022},
  volume = {317},
  pages = {115369},
  doi = {10.1016/j.jenvman.2022.115369},
}

@ARTICLE{Sun2026TII,
  author = {Sun, Kexin and Zhao, Zhiheng and Yang, Hongxia and Zhang, Jie and Huang, George Q.},
  title = {Curriculum Engineering: Structured Learning for Large Language Models ({LLMs}) Through Curriculum Based Retrieval},
  journal = {IEEE Transactions on Industrial Informatics},
  year = {2026},
  volume = {22},
  number = {1},
  pages = {555--566},
  doi = {10.1109/TII.2025.3610129},
}

@ARTICLE{Xu2023Vulnerability,
  author = {Xu, Senrong and Li, Liangyue and Li, Zenan and Yao, Yuan and Xu, Feng and Chen, Zulong and Lu, Quan and Tong, Hanghang},
  title = {On the Vulnerability of Graph Learning-Based Collaborative Filtering},
  journal = {ACM Transactions on Information Systems},
  year = {2023},
  volume = {41},
  number = {4},
  pages = {Article 87},
  doi = {10.1145/3572834},
}

@ARTICLE{Zhang2024NetworkedMultiagent,
  author = {Zhang, J. and Sang, L. and Xu, Y. and Sun, H.},
  title = {Networked Multiagent-Based Safe Reinforcement Learning for Low-Carbon Demand Management in Distribution Networks},
  journal = {IEEE Transactions on Sustainable Energy},
  year = {2024},
  volume = {15},
  number = {3},
  pages = {1528--1545},
  month = jul,
  doi = {10.1109/TSTE.2024.3355123},
}
\end{document}